\documentclass{article}
\usepackage[utf8]{inputenc}
\usepackage[T1]{fontenc}
\usepackage{graphicx}
\usepackage{subcaption}
\usepackage{xcolor}
\usepackage{booktabs}
\usepackage{authblk}
\usepackage{dirtree}
\usepackage{pifont}

\usepackage{xurl}
\usepackage{dirtree}
\usepackage{fontawesome5}
\usepackage{dirtree}
\usepackage{fontawesome5}
\usepackage[most]{tcolorbox}
\usepackage{xcolor}

\usepackage{fancyvrb}
\usepackage{verbatim}

\usepackage[super,comma,sort&compress]{natbib}

\makeatletter
\renewcommand\AB@authnote[1]{}
\renewcommand\AB@affilnote[1]{}
\makeatother
\DeclareUnicodeCharacter{2212}{\textminus}
\usepackage{booktabs}
\usepackage{multirow}
\usepackage{array}
\usepackage{graphicx}
\renewcommand{\arraystretch}{1.4} % improves row spacing
\usepackage{float} % for [H] placement
\usepackage{tabularx}
\usepackage{ragged2e} % for \RaggedRight inside cells
\usepackage{booktabs} % For professional-looking tables (\toprule, \midrule, \bottomrule)
\usepackage{multirow} % To span rows in a table
\usepackage{siunitx} % For aligning numbers by decimal point
\usepackage[left=2cm, right=2cm, top=2.25cm, bottom=2.25cm, headheight=12pt, letterpaper]{geometry}
\usepackage[labelfont={bf,sf}, labelsep=period, justification=raggedright]{caption}
\usepackage[colorlinks=true, allcolors=blue]{hyperref}

\title{\Large Cataract-LMM Large-Scale Multi-Source Multi-Task Benchmark for Deep Learning in Surgical Video Analysis}

\author[(1)]{Mohammad Javad Ahmadi$^{(1)}$}
\author[(1)]{Iman Gandomi$^{(1)}$}
\author[(2)]{Parisa Abdi$^{(2)}$}
\author[(2)]{Seyed-Farzad Mohammadi$^{(2)}$}
\author[(1)]{Amirhossein Taslimi$^{(1)}$}
\author[(2)]{Mehdi Khodaparast$^{(2)}$}
\author[(3)]{Hassan Hashemi$^{(3)}$}
\author[(4)]{Mahdi Tavakoli$^{(4)}$}
\author[(1)]{Hamid D. Taghirad$^{(1)}$}

\affil[(1)]{$^{(1)}$Applied Robotics and AI Solutions (ARAS), Faculties of Electrical and Computer Engineering, K.N. Toosi University of Technology, Tehran, Iran}
\affil[(2)]{$^{(2)}$Translational Ophthalmology Research Center, Farabi Eye Hospital, Tehran University of Medical Sciences, Tehran, Iran}
\affil[(3)]{$^{(3)}$Noor Ophthalmology Research Center, Noor Eye Hospital, Tehran University of Medical Sciences, Tehran, Iran}
\affil[(4)]{$^{(4)}$Departments of Electrical and Computer Engineering \& Biomedical Engineering, University of Alberta, Edmonton, AB, Canada}
\date{\vspace{2mm}{\footnotesize \textbf{Correspondence}: Hamid D. Taghirad (\href{mailto:taghirad@kntu.ac.ir}{taghirad@kntu.ac.ir}) \quad \textbf{Co-correspondence}: Parisa Abdi (\href{mailto:pabdi@sina.tums.ac.ir}{pabdi@sina.tums.ac.ir})}}
\begin{document}
\maketitle

\begin{abstract}
Computer-assisted surgery research requires large, deeply annotated video datasets that capture clinical and technical variability. Existing cataract surgery resources lack the diversity and annotation depth required to train generalizable deep-learning models. To address this gap, we present a dataset of 3,000 phacoemulsification cataract surgery videos acquired at two surgical centers from surgeons with varying expertise. The dataset provides four annotation layers: temporal surgical phases, instance segmentation of instruments and anatomical structures, instrument–tissue interaction tracking, and quantitative skill scores based on competency rubrics adapted from ICO-OSCAR and GRASIS. We demonstrate the technical utility of the dataset through benchmarking deep learning models across four tasks: workflow recognition, scene segmentation, instrument–tissue interaction tracking, and automated skill assessment. Furthermore, we establish a domain-adaptation baseline for phase recognition and instance segmentation by training on one surgical center and evaluating on a held-out center. Ultimately, these multi-source acquisitions, multi-layer annotations, and paired skill–kinematic labels facilitate the development of generalizable multi-task models for surgical workflow analysis, scene understanding, and competency-based training research.
\end{abstract}

\section*{Background \& Summary}
The persistent gap between growing global surgical demand and the capacity of the trained surgical workforce \cite{1} highlights the need to develop scalable solutions that can enhance training paradigms and optimize workflow management \cite{2}. Computer-assisted surgery (CAS) systems are one approach to address this challenge, with applications in preoperative planning \cite{3}, intraoperative guidance \cite{4}, and standardized postoperative assessment \cite{5,6}. The development and validation of these advanced CAS capabilities fundamentally depend on access to large-scale, deeply annotated surgical video datasets that capture procedural phases, instrument-tissue interactions, and technical skill cues \cite{7,8}.

Phacoemulsification cataract surgery is the most common ophthalmic procedure worldwide and the primary intervention for avoidable blindness \cite{9,10}. This makes it a critical domain for developing data-driven CAS with potential applications in clinical workflows and training \cite{11,12}. 

The domain of CAS in cataract surgery advanced through the introduction of some key benchmarks. The CATARACTS challenge \cite{13} established the primary baseline for phacoemulsification, providing tool-presence and workflow annotations for 50 videos. Its derivative, CaDIS \cite{14}, utilized a subset of these videos to introduce pixel-wise semantic segmentation. More recently, benchmarks such as CatRel \cite{15}, SICS-105 \cite{16}, and Sankara-MSICS \cite{17} have introduced phase and segmentation labels for phacoemulsification and manual small-incision cataract surgery (MSICS). At a larger scale, Cataract-1K \cite{18} released 1,000 phacoemulsification procedures with phase and segmentation annotations for a subset of videos.

Despite these contributions, a significant gap remains in integrating these tasks into a unified, clinically representative resource. Existing datasets are primarily restricted to single-center data. Furthermore, they often fail to capture a broad spectrum of surgical expertise or a variety of complex events beyond ideal scenarios. Crucially, these resources lack the specific linkage between dense instrument tracking and objective skill assessment based on blind scoring across distinct indicators, which is required to accurately model proficiency.

To address this gap, we present the Cataract-LMM (Large-scale, Multi-source, Multi-task) Dataset, a dataset of 3,000 phacoemulsification procedures recorded at two distinct clinical centers (Farabi and Noor Eye Hospitals, Tehran, Iran) between December 2021 and March 2025. The dataset is enriched with four complementary layers of annotations on subsets of the data:
\begin{enumerate}
    \item {Temporal Phase Labels (Phase):} Frame-wise annotations for 13 surgical phases across 150 videos to support automated workflow recognition.
    \item {Instance Segmentation Masks (Segmentation):} Pixel-wise masks for 10 instruments and 2 tissue classes in 6,094 frames from 150 videos to enable detailed scene parsing.
    \item {Spatiotemporal Interaction Masks (Tracking):} Frame-by-frame segmentation and tracking of instrument–tissue interactions in 170 videos for modeling surgical dynamics.
    \item {Quantitative Skill Assessment (Skill):} Objective skill scores for 170 videos using a systematic, multi-criteria rubric, providing a foundation for standardized skill assessment.
\end{enumerate}

By incorporating multiple annotations and including surgeons with varying experience levels across two centers, this dataset provides the procedural and technical diversity required to benchmark and develop multi-task domain-adaptive CAS models.

\section*{Methods} \label{Methods}

\subsection*{Ethical Approval}
This study was conducted in accordance with the Declaration of Helsinki and received ethical approval from the Tehran University of Medical Sciences (IR.TUMS.FARABIH.REC.1400.063), and the National Institute for Medical Research Development (IR.NIMAD.REC.1401.023). 
Written informed consent for participation and for the use and sharing of surgical video data for research purposes was obtained from all participants at the time of data collection. 
All data were fully de-identified prior to analysis to protect patient and surgeon privacy.

\subsection*{Data Acquisition and Curation}
A total of 3,000 phacoemulsification cataract surgery videos were prospectively collected between December 2021 and March 2025 from two ophthalmology centers in Tehran, Iran: Farabi Eye Hospital and Noor Eye Hospital. The acquisition strategy was intentionally multi-source, designed to capture procedural and technical variability. Procedural variability was introduced by including surgeons with a range of experience levels, with videos contributed by residents, fellows, and expert attendings. Technical variability was introduced by using two distinct, microscope-mounted camera setups: a Haag-Streit HS Hi-R NEO 900 (recording at 720×480 resolution and 30 fps) at Farabi Hospital, and a ZEISS ARTEVO 800 digital microscope (recording at 1920×1080 resolution and 60 fps) at Noor Hospital.

Video files were saved without post-processing and curated through a two-stage process. First, a technical quality screen was performed to exclude recordings based on pre-defined criteria: incomplete procedures, poor focus, or excessive glare obscuring key anatomical structures. Second, the remaining videos underwent the de-identification process. This resulted in a final curated dataset of 3,000 procedures, comprising 2,930 from Farabi Hospital and 70 from Noor Hospital, with a total video duration of 1,134.2 hours.

\subsection*{Annotation Protocols}
The Cataract-LMM dataset provides four comprehensive annotation layers across overlapping subsets to support a wide range of advanced surgical and AI research. It offers significant advantages over existing resources in terms of scale, multi-source diversity, and the depth of its multi-layered annotations, as detailed in the comparative analysis in Table 1. 

To select videos for the annotated subsets from the larger corpus of 3,000 procedures, the baseline cases were chosen randomly. However, we also employed a content-aware, purposeful sampling strategy to incorporate targeted cases designed to maximize the informational value and generalization capability of the benchmarks. Unlike relying solely on simple random sampling, which risks over-representing routine cases, this protocol incorporated a purposeful sampling component to additionally target hard negatives and edge cases. The selection process was conducted on the fully de-identified dataset under the supervision of a senior ophthalmic surgeon (P.A.) to mitigate selection bias while ensuring clinical representativeness.

The sampling focused on three specific dimensions: (1) {Procedural Heterogeneity}, ensuring the capture of stochastic workflow variations and intra-operative events (evident in the temporal variance shown in Figure 6); (2) {Visual Complexity}, actively incorporating frames with significant imaging artifacts such as specular reflections and occlusions (Figure 3); and (3) {Behavioral Diversity}, ensuring a balanced distribution of surgical proficiency from novice to expert, as evidenced by the resulting comprehensive score distribution (Figure 11) and the distinct motion patterns (Figure 13 and Figure 14).
        
Detailed methodologies for each annotation protocol are presented in the following sections.

\begin{table}[htbp]
\centering
\caption{Comparison of Cataract-LMM with publicly available cataract surgery datasets.}
\label{table1}
\scriptsize % Used to fit the large number of columns
\renewcommand{\arraystretch}{1.4}
\setlength{\tabcolsep}{2pt}
\resizebox{\textwidth}{!}{%
\begin{tabular}{
  l % Feature Name
  p{2.7cm} % CATARACTS
  p{1.7cm} % CaDIS
  p{1.7cm} % CatRel
  p{1.9cm} % SICS-105
  p{2.9cm} % Cat-MSICS
  p{2.3cm} % Cat-1K
  p{2.3cm} % LMM
}
\toprule
 \textbf{Feature / Dataset} & \textbf{CATARACTS \cite{13}} & \textbf{CaDIS \cite{14}} & \textbf{CatRel \cite{15}} & \textbf{SICS-105 \cite{16}} & \textbf{Sankara-MSICS \cite{17}} & \textbf{Cataract-1K \cite{18}} & \textbf{Cataract-LMM} \\
\midrule
 \textbf{Year} & 2017 & 2019 & 2020 & 2025 & 2025 & 2021--2023 & \textbf{2021--2025} \\
 \cmidrule(l){1-8}
 \textbf{Total Cases} & 50 & 25 & 22 & 105 & 53 & 1,000 & \textbf{3,000} \\
 \cmidrule(l){1-8}
 \textbf{Source} & Single & Single & Single & Single & Single & Single & \textbf{Multi} \\
 \cmidrule(l){1-8}
 \textbf{Resolution} & 1920$\times$1080 & 960$\times$540 & 224$\times$224 & 1920$\times$1080 & 1024$\times$768 & 1024$\times$768 &  \begin{tabular}[t]{@{}l@{}}\textbf{720$\times$480}\\\textbf{1920$\times$1080}\end{tabular} \\
 \cmidrule(l){1-8}
 \textbf{FPS} & 30 & Not Available & Not Available & 30 & 30 & 30 & \textbf{30 \& 60} \\
 \cmidrule(l){1-8}
 \textbf{Phase Recognition} & 50 videos \newline (14 phases) & Not Available & 22 videos \newline (2 phases) & 105 videos \newline (13 phases) & 53 videos \newline (18 phases) & 56 videos \newline (13 phases) & \textbf{150 videos \newline (13 phases)} \\
 \cmidrule(l){1-8}
 \textbf{Tool Presence} & 50 videos & 25 videos & 11 videos & Not Available & 53 videos & 30 videos & \textbf{150 videos} \\
 \cmidrule(l){1-8}
 \textbf{Instance Segmentation} & Not Available & \begin{tabular}[t]{@{}l@{}}4,670 frames\\(25 videos)\end{tabular} & \begin{tabular}[t]{@{}l@{}}216 frames\\(11 videos)\end{tabular} & Not Available & \begin{tabular}[t]{@{}l@{}}3,527 frames\\(53 videos)\end{tabular} & \begin{tabular}[t]{@{}l@{}}2,256 frames\\(30 videos)\end{tabular} & \textbf{\begin{tabular}[t]{@{}l@{}}6,094 frames\\(150 videos)\end{tabular}} \\
 \cmidrule(l){1-8}
 \textbf{Tracking} & Not Available & Not Available & Not Available & Not Available & Not Available & Not Available & \textbf{\begin{tabular}[t]{@{}l@{}}170 videos\\(469,118 frames)\end{tabular}} \\
 \cmidrule(l){1-8}
 \textbf{Skill Assessment} & Not Available & Not Available & Not Available & Not Available & Not Available & Not Available & \textbf{\begin{tabular}[t]{@{}l@{}}170 videos\\(1--5 Scale Rubric)\end{tabular}} \\
\bottomrule
\end{tabular}%
}
\end{table}

\subsubsection*{Phase Recognition Dataset Annotation Protocol}
A subset of 150 videos (129 from Farabi Hospital, 21 from Noor Hospital), with a total duration of 28.55 hours, was annotated with temporal phase labels to facilitate automated surgical workflow analysis and video-preprocessing pipelines \cite{19}.

To create a standardized annotation framework, a taxonomy of 13 distinct surgical phases was defined based on the established procedural steps in phacoemulsification cataract surgery \cite{18}. This taxonomy covers the entire procedure from Incision to Tonifying-Antibiotics, including an Idle phase to label surgical inactivity or instrument exchange. Representative frames illustrating the visual characteristics of each phase from both hospital sources are presented in Figure 1.
\begin{figure} [htbp]
    \centering
    \includegraphics[width=0.75\linewidth]{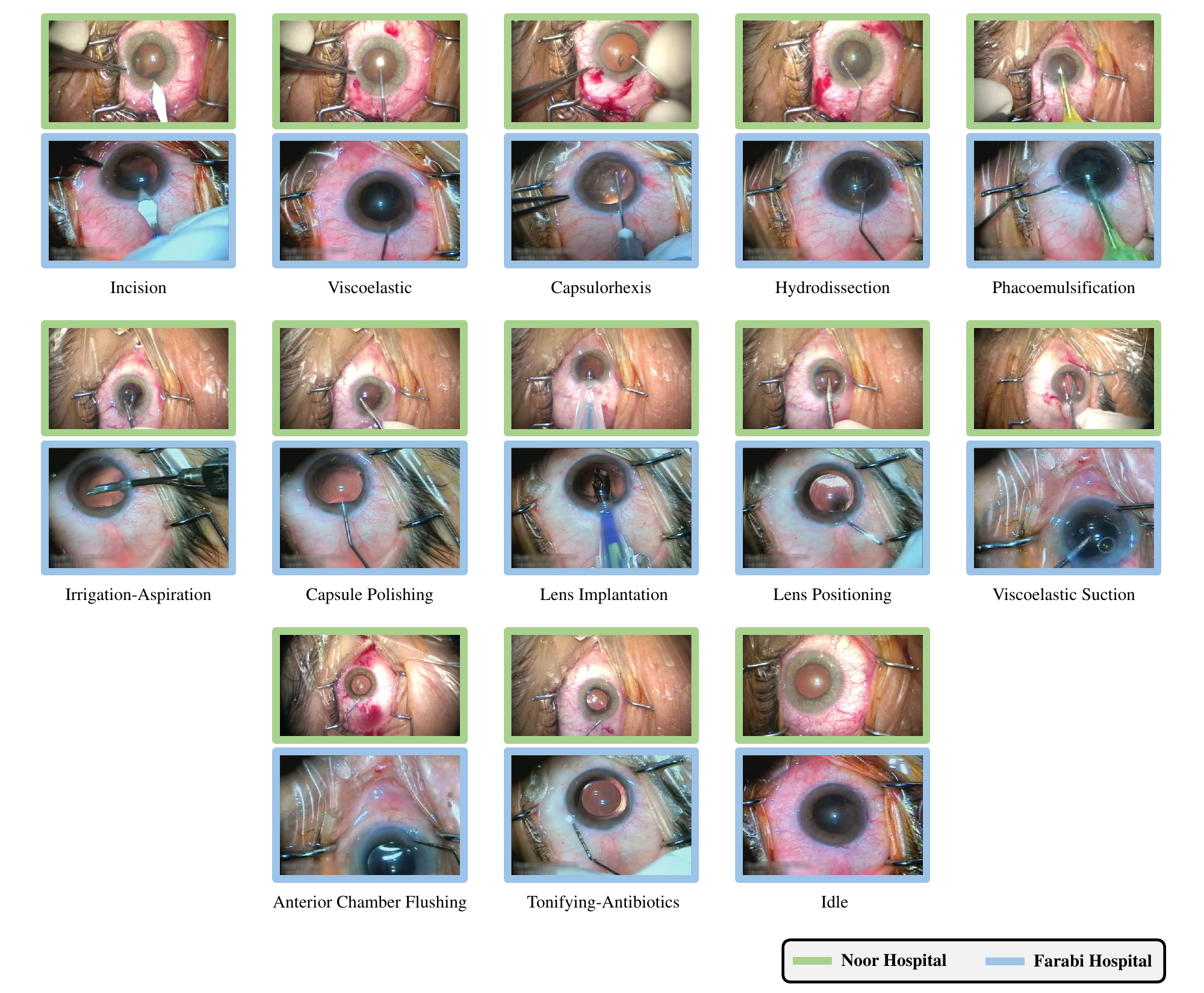}
    \caption{Visual overview of key surgical phases from both clinical centers, illustrating domain shift.}
    \label{figure1}
\end{figure}

To ensure the generation of a highly reliable and reproducible ground truth, the phase annotation process was executed as a multi-stage scientific protocol by a workforce of three ophthalmology residents (Years 2–4). Prior to the primary data collection, annotators underwent a comprehensive workshop to standardize their interpretation of the 13-phase taxonomy. This training established video anchors, gold-standard clips defining precise visual cues for phase boundaries, and included joint annotation sessions to align the workforce on edge cases. The workforce's proficiency was quantitatively validated using a designated calibration set of the 18 videos, partitioned into a Pilot Set ($n=5$) for initial alignment and a held-out Validation Set ($n=13$). The annotators achieved a Global Fleiss’ Kappa of $\kappa=0.924$ on the validation subset, confirming high inter-rater reliability before proceeding to the full dataset.

Annotation was performed using our custom-developed platform (SurgiNote) via a "coarse-to-fine" protocol, where annotators first localized temporal transitions at the second-level temporal resolution before refining the boundaries frame-by-frame. To maintain longitudinal consistency, a {hybrid expert review protocol} was implemented. This protocol, conducted under the supervision of two senior reviewers, a senior expert in AI (M.J.A.) and an associate professor and attending surgeon experienced as a surgical trainer (P.A.), consisted of a full audit of $\sim$10\% of the videos and a targeted review of transition points for the remaining 90\%. Discrepancies identified during expert review were adjudicated in weekly dispute resolution sessions, where final labels were determined through rubric-referenced group consensus to mitigate individual observer bias.

\subsubsection*{Instance Segmentation Dataset Annotation Protocol}
To enable detailed surgical scene analysis, an instance segmentation subset was created from 6,094 frames sampled from the 150 videos, comprising 3,932 frames from Farabi and 2,162 from Noor. Frames were annotated with instance-level segmentation masks for 12 classes: two ocular structures (Pupil, Cornea) and ten surgical instruments (Primary knife, Secondary knife, Capsulorhexis Cystotome, Capsulorhexis Forceps, Phaco Handpiece, I/A Handpiece, Second Instrument, Forceps, Cannula, and Lens Injector). Representative examples of these instruments from each hospital source, highlighting the inherent domain shift, are illustrated in Figure 2.
\begin{figure}[htbp]
    \centering
    \includegraphics[width=0.95\linewidth]{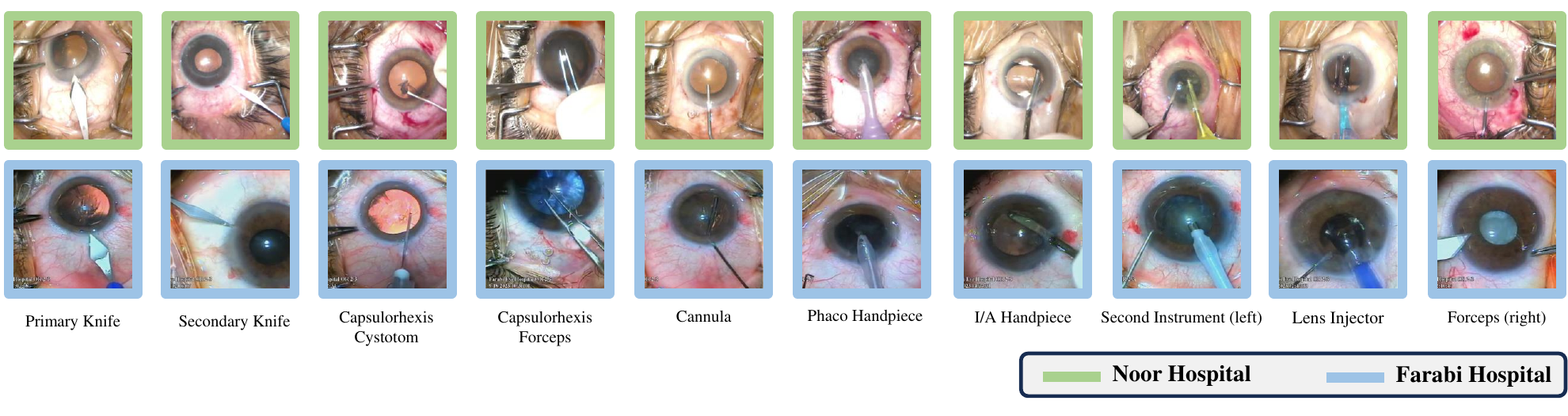}
    \caption{Examples of surgical instruments from the two data sources, illustrating domain shift.}
    \label{figure2}
\end{figure}

To curate this subset, a systematic sampling methodology was used to create a diverse and challenging instance segmentation dataset. 
Frames were randomly sampled from the 150 videos, covering all 13 surgical phases, every surgical instrument utilized, and the relevant anatomical structures. To maximize temporal diversity and avoid near-duplicate frames, a minimum interval of 0.5 seconds was enforced between any two frames sampled from the same video. 

Beyond maximizing temporal diversity, the selection process intentionally incorporated frames depicting common visual difficulties to create a challenging and realistic benchmark, while frames with severe, non-informative motion blur or occlusion were excluded. As a result, Figure 3 illustrates these visual difficulties with representative frames and their corresponding segmentation masks, including examples of high inter-instrument similarity, boundary ambiguity from motion or depth of field, and specular reflections. 
\begin{figure}[htbp]
    \centering
    \includegraphics[width=0.85\linewidth]{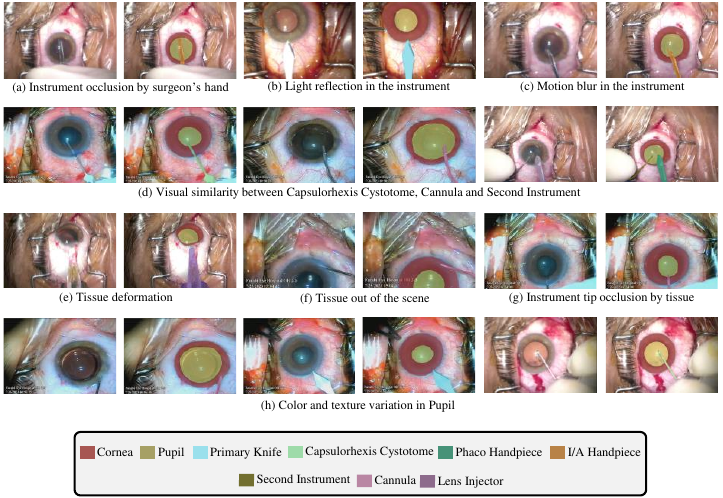}
    \caption{Examples of common visual challenges for instance segmentation in the dataset.}
    \label{figure3}
\end{figure}

To ensure the generation of high-fidelity ground truth masks, the annotation process was conducted using the Roboflow computer vision collaboration platform by a workforce of eight ophthalmology residents (Years 2–4). Prior to annotation, a visual ontology was defined to standardize polygon tightness for challenging classes, specifically addressing boundary ambiguities arising from tissue transparency (Cornea) and complex visual obstructions caused by cast shadows and instrument occlusions. The workforce was qualified on a calibration set, achieving a mean Intersection over Union (mIoU) of 0.874 and a semantic class accuracy of 0.992 on the blind validation subset ($n=250$), ensuring both spatial precision and expert-level object identification. \\
Following successful workforce calibration, the primary annotation of the 6,094 frames (samples) was performed using detailed polygon masks. To maintain quality control, a random sampling audit protocol was implemented within Roboflow, where 15\% of annotations were reviewed by two senior reviewers: a senior expert in AI (M.J.A.) and an Associate Professor and surgeon trainer (P.A.). Frames failing strict boundary adherence or semantic classification checks were rejected and returned for correction, ensuring the final dataset meets the defined quality criteria.

\subsubsection*{Object Tracking Dataset Annotation Protocol}
To enable the quantitative analysis of the spatiotemporal dynamics of surgical technique, a tracking dataset was created from 170 video clips of the capsulorhexis phase. Proficiency in this phase is highly correlated with overall procedural success and patient outcomes \cite{20}. 

This subset was curated independently from the phase recognition videos by screening the full raw dataset ($n=3,000$) to identify clips exhibiting a wide spectrum of psychomotor proficiency, ensuring a robust distribution of skill levels for kinematic analysis.

Due to the large scale of the tracking dataset, manual annotation was augmented by a verified {Human-in-the-Loop} semi-automated pipeline designed to ensure both scalability and medical fidelity. The workflow commenced with an AI-based pre-annotation stage, where the Ultralytics YOLO11-L model, fine-tuned on the validated instance segmentation subset, generated initial frame-wise masks. Temporal consistency was enforced using a custom tracking logic based on the BoT-SORT \cite{21} algorithm, which utilized a tracklet memory buffer and Kalman filtering to bridge occlusion gaps and maintain persistent instance identifiers. Functional keypoints, specifically instrument tips, were computed algorithmically via a constraint-based geometric regression script to identify the distal tip. This method applies least-squares line fitting to the mask body to identify the distal tip, reducing the variability inherent in manual pixel selection.

The algorithmically generated data subsequently underwent a human verification phase. We deployed the same workforce of eight ophthalmology residents qualified for the instance segmentation task to review the data within a hybrid annotation environment. The correction protocol focused on two critical quality dimensions: spatial boundary adherence, where annotators refined mask vertices to strictly encapsulate instruments during rapid motion, and identity verification, where any algorithmic ID switches were corrected to ensure unbroken trajectories. To quantify the reliability of this pipeline, we conducted a blinded inter-rater reliability study on a stratified validation subset of 10 video clips. The analysis yielded a near-perfect Association Accuracy (AssA) of 98.4, Detection Accuracy (DetA) of 82.7, and a Higher Order Tracking Accuracy (HOTA) of 90.2, indicating spatial precision, temporal coherence, and reproducibility of the annotations.
        
This process yielded a rich set of multi-layered annotations for each frame in the video clips, as detailed in Table 2. A representative frame with its corresponding multi-layered annotations is shown in Figure 4. The tracking annotations are designed to enable the extraction of surgeons’ motion information and to characterize instrument-tissue interaction patterns. By linking keypoints and persistent identifiers over time, two-dimensional motion trajectories and kinematic descriptors such as path length, velocity, and jerk can be computed. Additionally, visual instrument-tissue intersection events (defined as mask overlaps in the frame plane), proximity to anatomical boundaries, and instrument utilization patterns can be quantified. As this subset is linked to expert skill ratings, these motion-derived metrics can be associated with proficiency to support objective performance assessment and the visualization of surgical motion paths.
\begin{table}[htbp]
    \centering
        \caption{Structure of the multi-layered annotations provided in the tracking dataset.}
    \label{table2}
\resizebox{0.85\textwidth}{!}{%
\begin{tabular}{|>{\raggedright\arraybackslash}p{3.5cm}|>{\raggedright\arraybackslash}p{4cm}|>{\raggedright\arraybackslash}p{8cm}|}
\hline
\textbf{Annotation Layer} & \textbf{Format} & \textbf{Description} \\
\hline
Instance Segmentation & Standard format (COCO) & Pixel-level masks identifying the boundaries of key surgical instruments and anatomical structures (pupil, cornea). \\
\hline
Bounding Boxes & \texttt{[x, \quad y, \quad width, \quad height]} & The coordinates of the tightest bounding box enclosing each segmented instance, with \texttt{x, \quad y} defining the top-left corner. \\
\hline
Persistent Instance IDs & Integer (\texttt{\textit{category\_id}}) & A unique integer identifier assigned to each distinct object instance (e.g., a specific forceps) that remains constant for that object throughout the entire video clip, enabling robust tracking. \\
\hline
Functional Keypoints & \texttt{[x, \quad y]} coordinates & Labeled coordinates for functionally critical points. This includes the instrument tip (defined as the distal-most functional point) and the geometric centroids of the cornea and pupil masks. \\
\hline
Motion Trajectories & Sequence of \texttt{[x, \quad y]} per \texttt{\textit{category\_id}} & A time-series of keypoint coordinates for each tracked object. This raw data enables the derivation of kinematic metrics, including the velocity, acceleration, and jerk of instrument and tissue movements. \\
\hline
\end{tabular}%
}
\end{table}
\begin{figure}[htbp]
    \centering
    \includegraphics[width=0.85\linewidth]{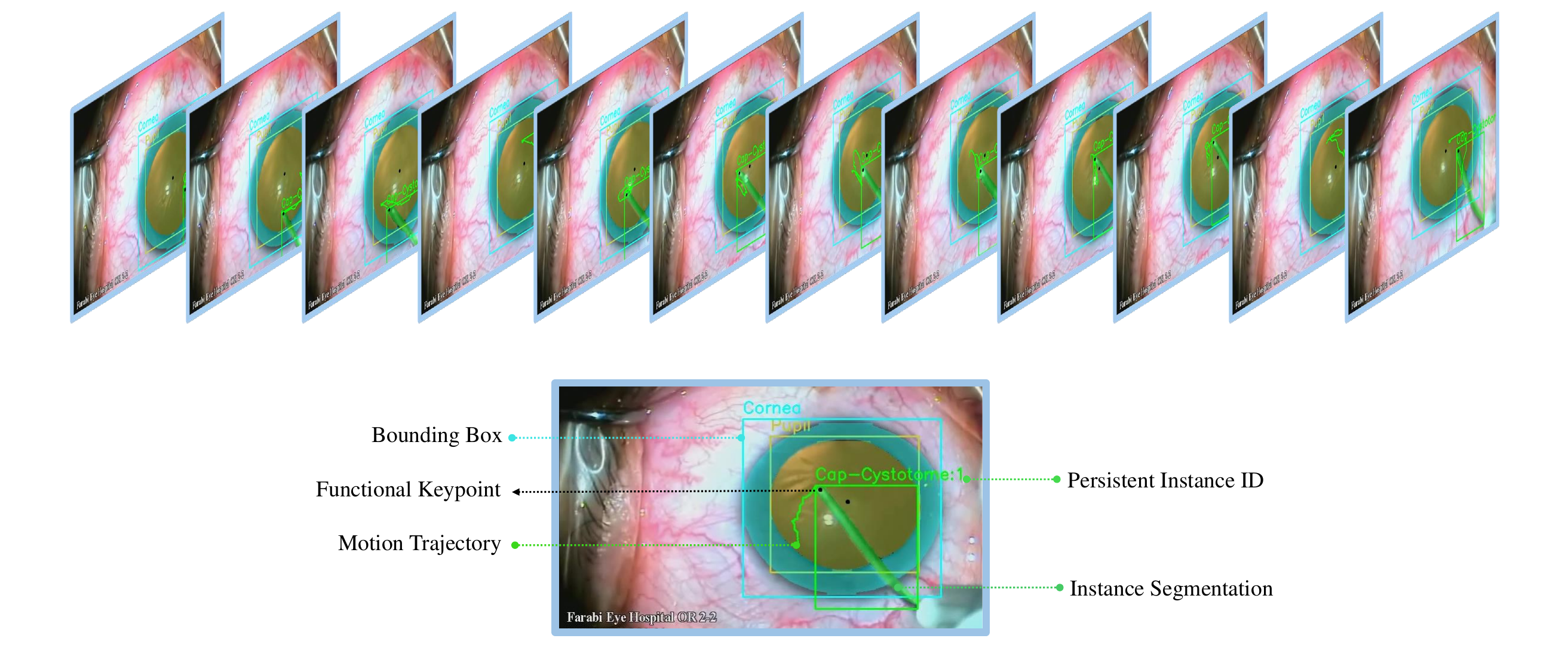}
    \caption{Example of multi-layered annotations for a single frame from the tracking dataset.}
    \label{figure4}
\end{figure}

\subsubsection*{Skill Assessment Dataset Description} \label{Skill Assessment Dataset Description}

To support competency-based training research and the development of automated feedback systems, the same 170 capsulorhexis video clips used for tracking were annotated with objective surgical skill scores. This linkage allows for the investigation of how expert-rated proficiency correlates with quantitative surgical motion information derived from instrument-tissue dynamics and trajectories. 

A video-based rubric was developed through a formal consensus process involving three consultant ophthalmic surgeons and two medical education experts. The panel adapted six performance indicators from validated standards (GRASIS \cite{22}, and ICO-OSCAR \cite{23}) that could be reliably assessed from video alone. 

Table 3 details this 6-indicator rubric, providing descriptive anchors for the 5-point rating scale for each indicator. The presence of critical Adverse Events was documented as a binary flag for each clip. Furthermore, to provide granular context for safety analysis, annotators recorded specific descriptions of these events (e.g., radial capsular tears, zonular dehiscence, small/large capsulorhexis, bleeding events) and relevant intraoperative risk factors (e.g., miotic pupil, mature cataract, corneal opacity) in a dedicated qualitative \texttt{comments} column included in the dataset.
\begin{table}[t]
    \centering
        \caption{The 6-indicator rubric used for skill assessment of the capsulorhexis video clips.}
    \label{table3}
\resizebox{0.9\columnwidth}{!}{%
\begin{tabularx}{\textwidth}{|>{\RaggedRight\arraybackslash}p{3cm}|%
                            >{\RaggedRight\arraybackslash}p{2.5cm}|%
                            >{\RaggedRight\arraybackslash}X|%
                            >{\RaggedRight\arraybackslash}X|%
                            >{\RaggedRight\arraybackslash}X|}
\hline
\textbf{Indicator} & \textbf{Source} & \textbf{Novice (Score 1–2)} & \textbf{Intermediate (Score 3–4)} & \textbf{Competent (Score 5)} \\
\hline
Instrument Handling & GRASIS \cite{22} &
Repeated, abrupt, or harsh movements; endless entry and exit. &
Selected or occasional inappropriate movements. &
Fine and smooth movements with no inappropriate actions. \\
\hline
Motion & ICO-OSCAR \cite{23} &
Unsure surgical plan with needless, in-doubt movements. &
A certain surgical plan with occasional unnecessary movements. &
Maximum effective movements; no unnecessary actions. \\
\hline
Tissue Handling & GRASIS \cite{22} &
Unnecessary force applied; damage to the cornea or conjunctiva. &
Suitable tissue interactions with minor, unintentional tissue damage. &
Excellent tissue interactions with no iatrogenic damage. \\
\hline
Microscope Use & GRASIS \cite{22} &
Multiple recentering and refocusing attempts are required. &
Few attempts to recenter or refocus. &
Eye kept centered with a good, focused view throughout. \\
\hline
Commencement of Flap & ICO-OSCAR \cite{23} &
Tentative chasing rather than controlled creation; numerous cortex disruptions. &
Flap pulled up after 2–3 tries; subtle cortex disruptions. &
Delicate and controlled approach; no cortex disruption. \\
\hline
Circular Completion & ICO-OSCAR \cite{23} &
Unable to achieve a circular rhexis; extension into the periphery. &
Difficulty achieving a continuous circular rhexis. &
Rapid, unaided, and controlled completion of the rhexis. \\
\hline
\end{tabularx}
}
\end{table}

A three-stage methodology was implemented to ensure reliable and reproducible skill assessments. Prior to the consensus phase, the inter-rater reliability of the initial blind ratings was evaluated using the Intraclass Correlation Coefficient (ICC). The analysis yielded an overall ICC of 0.87, indicating excellent agreement among the three independent raters. The final scoring process followed these steps:
\begin{enumerate}
    \item {Double-blind Tri-rating:} Three board-certified ophthalmic surgeons independently scored each clip without knowledge of the surgeon’s identity or their peers’ ratings.
    \item {Supervisor Adjudication:} A senior consultant reviewed all ratings. Any disagreement between raters exceeding one point on the 5-point scale for any indicator triggered a consensus discussion to resolve the discrepancy and assign a final score.
    \item {Score Aggregation:} For performance indicators where the inter-rater disagreement was within the tolerance threshold ($\le 1$ point), the final indicator score was calculated as the arithmetic mean of the three independent ratings. In cases requiring adjudication, the consensus score was utilized. Finally, an overall proficiency score for the clip was computed as the unweighted mean of the six final indicator scores.
\end{enumerate}

The resulting score distribution and construct validity of the rubric are characterized in the Technical Validation section.

\subsection*{Experiments Methodology}
This section details the technical validation protocols for surgical phase recognition, instance segmentation, object tracking, and objective skill assessment, including the model architectures, training configurations, and evaluation metrics used to establish performance baselines for each task.

\subsubsection*{Experimental Methodology for Phase Recognition}
To demonstrate the dataset's utility, we established phase recognition baselines evaluated at both the clip level and video level. Clip-level (causal) models simulate real-time intraoperative deployment, whereas video-level (non-causal) models are appropriate for post-hoc video indexing and retrospective analysis. We employed both two-stage and end-to-end deep learning strategies and explicitly measured the models' robustness to domain shift.

For feature extraction in the two-stage framework, we utilized Convolutional Neural Network (CNN) backbones (ResNet50, EfficientNet-B5) pre-trained on ImageNet, as well as foundation models (DINO, CLIP) using their frozen ViT-B/16 image encoders to extract frame-level spatial features.

For the clip-level benchmark, feature sequences were modeled using Recurrent Neural Networks (Long Short-Term Memory network (LSTM) \cite{24} and Gated Recurrent Unit (GRU) \cite{25}) on short 10-frame clips. We also benchmarked end-to-end video recognition models pre-trained on Kinetics-400, including 3D-CNNs (SlowFast \cite{26}, X3D \cite{27}, R(2+1)D \cite{28}, MC3 \cite{28}, R3D \cite{29}) and Vision Transformers (MViT \cite{30}, Video Swin Transformer \cite{31}). For the video-level benchmark, which processes full procedural sequences, we employed stronger temporal architectures: the multi-stage temporal convolutional network (MS-TCN) framework (specifically, the TeCNO implementation) \cite{32} and the transformer-based ASFormer \cite{33}.

To assess model generalization, we partitioned the dataset based on the clinic of origin. The training set (80 videos) and validation set (26 videos) were drawn exclusively from the Farabi hospital. The test set consisted of 44 videos: 23 unseen videos from the Farabi hospital (in-distribution) and all 21 videos from the Noor hospital (out-of-distribution). This strategy directly evaluates the models' ability to generalize to data from a different clinical setting.

All videos were downsampled to 4 frames per second (fps), as initial validation showed this rate offered a favorable balance between model performance and computational cost. Regarding feature preparation, for the CNN backbones (ResNet, EfficientNet), the models were initially fine-tuned for frame-wise classification using a two-layer Multi-Layer Perceptron (MLP) head, after which the CNN weights were frozen. In contrast, for DINO and CLIP, we utilized the pre-trained weights directly without further fine-tuning. The temporal models were subsequently trained on the sequences of extracted features.

To handle challenging classes, the visually similar and underrepresented phases, Viscoelastic and Anterior Chamber Flushing, were merged into a single class.  To mitigate the natural class imbalance, we implemented a hybrid sampling strategy during training. Clips from over-represented phases were randomly undersampled, while clips from under-represented phases were oversampled using random horizontal flipping and brightness adjustments. Key hyperparameters, detailed in Table 4, were kept consistent across all experiments.
\begin{table}[htbp]
\centering
\caption{Hyperparameter configuration for phase recognition experiments.}
\label{table4}
% \resizebox{0.8\columnwidth}{!}{%
\begin{tabular}{l c}
\hline
\textbf{Hyperparameter} & \textbf{Value} \\
\hline
Input Frame Resolution & $224 \times 224$ pixels \\
Batch Size & 32 \\
Optimizer & Adam \\
Learning Rate & $1 \times 10^{-4}$ \\
Weight Decay & $1 \times 10^{-3}$ \\
Dropout Rate & 0.5 \\
Temporal Model Hidden Dim. & 256 \\
MLP Hidden Layer & 128 \\
Number of ASFormer Decoders, Layers, Feature Maps & 2, 11, 64 \\
Number of MS-TCN (TeCNO) Stages, Layers, Feature Maps & 1, 6, 256 \\
\hline
\end{tabular}
% }
\end{table}

Model performance was evaluated using two sets of metrics. For clip-level models, we reported frame-level accuracy (the proportion of correctly classified frames), macro-averaged precision, recall, and F1 score. The macro-averaged metrics were chosen to ensure that each phase contributes equally to the aggregate score, providing a balanced assessment that is robust to the inherent class imbalance in the dataset. 

For video-level models, we additionally included temporal consistency metrics: the Edit Score and segmental F1 overlap scores at IoU thresholds of 10\%, 25\%, and 50\%. Standard frame-level metrics (like accuracy) treat frames independently and do not penalize 'flickering' or over-segmentation errors. We included the Edit Score and segmental F1 overlap to explicitly evaluate the temporal continuity and the quality of the predicted phase boundaries, ensuring the model captures the long-term structure of the surgical procedure.

\subsubsection*{Experimental Methodology for Instance Segmentation}
To demonstrate the utility of the instance segmentation dataset, we provide baseline performance benchmarks using established deep learning models. The experimental setup was designed to evaluate both supervised and zero-shot approaches and to assess model performance across varying levels of semantic granularity.

Three distinct tasks were defined by grouping the 12 base classes to address different potential use cases. Certain applications, such as distinguishing active surgical periods from idle time, only require detecting the presence of a generic instrument rather than its specific type. Accordingly, Task 1 merges all 10 instruments into a single “Instrument” class (3 classes total). Task 2 offers a more granular, balanced 9-class scheme by merging only the most visually and functionally similar instruments (e.g., “Primary Knife” and “Secondary Knife” become “Knife”). Finally, Task 3 provides the highest level of detail by treating all 12 classes as distinct. The precise class mappings for each task are detailed in Table 5.
\begin{table}[htbp]
\centering
\caption{Semantic class grouping strategy for the three defined instance segmentation tasks.}
\label{table5}
\begin{tabular}{llll}
\hline
\textbf{Base Class}           & \textbf{Task 1 Grouping} & \textbf{Task 2 Grouping}    & \textbf{Task 3 Grouping}      \\ \hline
Cornea                        & Cornea                   & Cornea                      & Cornea                        \\
Pupil                         & Pupil                    & Pupil                       & Pupil                         \\
Primary Knife                 & Instrument               & Knife                       & Primary Knife                 \\
Secondary Knife               & Instrument               & Knife                       & Secondary Knife               \\
Capsulorhexis Cystotome       & Instrument               & Instrument                  & Capsulorhexis Cystotome       \\
Second Instrument             & Instrument               & Instrument                  & Second Instrument             \\
Cannula                       & Instrument               & Instrument                  & Cannula                       \\
Capsulorhexis Forceps         & Instrument               & Capsulorhexis Forceps       & Capsulorhexis Forceps         \\
Forceps                       & Instrument               & Forceps                     & Forceps                       \\
Lens Injector                 & Instrument               & Lens Injector               & Lens Injector                 \\
Phaco Handpiece              & Instrument               & Phaco Handpiece            & Phaco Handpiece              \\
I/A Handpiece                 & Instrument               & I/A Handpiece               & I/A Handpiece                 \\ \hline
\textbf{Total Classes}        & \textbf{3}               & \textbf{9}                  & \textbf{12}                   \\ \hline
\end{tabular}
\end{table}

A suite of supervised models, all pre-trained on the COCO dataset \cite{34}, was selected for benchmarking. This included Mask R-CNN \cite{35} with a ResNet-50 backbone, alongside the YOLOv8-L and YOLOv11-L \cite{36} models. In parallel, the generalization capabilities of zero-shot models, specifically the Segment Anything Model (SAM) \cite{37} and SAM2 \cite{38}, were assessed without any fine-tuning.

The 6,094 annotated frames were split into training, validation, and test sets with a 70/20/10 ratio. This division was performed at the video level to ensure standardized benchmarking. To prevent data leakage and ensure the model's generalization ability, all frames from a single surgical video were assigned to only one of the three data splits.

All input images were resized to 640×640 pixels, and data augmentation strategies were applied, including random Gaussian blur, brightness adjustments, and hue-saturation-value (HSV) color space jittering. The AdamW optimizer was used for all supervised training. The specific hyperparameters for the primary models are detailed in Table 6.
\begin{table}[htbp]
\centering
\caption{Hyperparameter configurations for the primary supervised instance segmentation models.}
\label{table6}
\begin{tabular}{l c c}
\hline
\textbf{Hyperparameter} & \textbf{Mask R-CNN} & \textbf{YOLOv8-L / YOLOv11-L} \\ \hline
Learning Rate           & $5 \times 10^{-4}$  & $8 \times 10^{-4}$            \\
Weight Decay            & $5 \times 10^{-4}$  & 0                             \\
Training Epochs         & 20                  & 80                            \\
Batch Size              & 8                   & 16                            \\ \hline
\end{tabular}
\end{table}

For the zero-shot evaluation, SAM and SAM2 were prompted with ground-truth bounding boxes to generate segmentation masks. Using ground-truth bounding boxes gives SAM/SAM2 an oracle localization signal; observed performance, therefore, represents an upper bound for zero-shot segmentation quality. This bounding-box-prompting strategy was selected to specifically assess the models' segmentation capabilities on given regions of interest, independent of their object detection or localization performance.

Performance was evaluated using mean Average Precision for instance segmentation (mask mAP), calculated over Intersection over Union (IoU) thresholds from 0.50 to 0.95, following the standard COCO evaluation protocol.

\subsubsection*{Experimental Methodology for Object Tracking}
To validate the technical quality and clinical relevance of the tracking annotations, we designed a comprehensive two-part protocol to characterize the spatial distribution of the effective surgical workspace and evaluate the proximity profile of tool-tissue interactions.

To delineate the effective Region of Interest (ROI) of the capsulorhexis, frame-wise instrument tip coordinates were expressed in a pupil-centric reference frame by computing the relative displacement $(\Delta x, \Delta y)$ between the instrument tip and the pupil centroid for every annotated frame. This normalization mitigates inter-patient anatomical variability and camera pose differences across clinical centers. The resulting point cloud was used to fit a two-dimensional Gaussian Kernel Density Estimator (KDE) to model the spatial distribution of the instrument trajectories. This workspace analysis was conducted independently for the {Lower-Skilled} and {Higher-Skilled} cohorts defined in the Skill Assessment Dataset section, with cohort assignment propagated from the video level to the frame-level tracking records via a standardized identifier-matching procedure, enabling a direct comparison of novice and expert spatial envelopes.

To quantify the spatial distribution of tool-tissue interactions, we computed the frame-wise Euclidean distance between the instrument tip and the pupil centroid, $d = \sqrt{\Delta x^2 + \Delta y^2}$, across the complete subset. The resulting distance population was summarized using a normalized histogram overlaid with a kernel density curve and annotated with the mean and median, characterizing both the central tendency of the interaction zone and the tail behavior associated with instrument insertion and withdrawal events.

\subsubsection*{Experimental Methodology for Skill Assessment} \label{EAssessment}
To validate the skill assessment dataset, we established a technical validation protocol using three complementary approaches: a quantitative video-based classification benchmark, a qualitative analysis of instrument motion trajectories, and a quantitative kinematic analysis.

For the video-based classification benchmark, the objective was to train models to distinguish between surgeon skill levels. To create a well-defined binary classification task, the continuous overall skill scores for the 170 clips were partitioned using a K-Means clustering algorithm (K=2). This process resulted in a {Lower-Skilled} group (n=63, mean score = 3.12 $\pm$ 0.38) and a {Higher-Skilled} group (n=107, mean score = 4.24 $\pm$ 0.37).

While we established this binary setup as a baseline, the Cataract-LMM dataset retains its full continuous overall scores and six individual rubric indicators, thus supporting diverse benchmarking formulations by future researchers, such as (i) regression on continuous scores, (ii) multi-class skill categorization via expert-defined thresholds or data-driven clustering, and (iii) objective assessment using motion-derived kinematic features.

The 170 video clips were then split at the video level into training (70\%), validation (15\%), and test (15\%) sets, ensuring no procedural overlap between sets.

To establish a comprehensive baseline, we benchmarked models representing three dominant architectural paradigms for video analysis: (i) 3D-CNNs (X3D-M \cite{27}, SlowFast R50 \cite{26}, R(2+1)D-18 \cite{28}, and R3D-18 \cite{29}); (ii) hybrid CNN-RNN models (CNN-LSTM and CNN-GRU); and (iii) a Transformer-based model (TimeSformer \cite{39}).

Input data for all models consisted of 100-frame snippets sampled at 10 frames per second, with a 10-frame overlap between consecutive snippets of the train split. During inference, video-level predictions were generated by aggregating the snippet-level outputs. Specifically, we calculated the arithmetic mean of the predicted posterior probabilities across all 100-frame windows extracted from a test clip to produce a single, deterministic skill classification for the procedure.

All frames were resized to $224 \times 224$ pixels. Models were trained for 25 epochs using the AdamW optimizer and a cosine annealing learning rate schedule. Key hyperparameters are detailed in Table 7. The performance of all models was evaluated using accuracy, precision, recall, and macro-averaged F1-score.
\begin{table}[htbp]
\centering
\caption{Hyperparameter settings for the video-based skill assessment classification benchmark.}
\label{table7}
\begin{tabular}{l l}
\hline
\textbf{Hyperparameter} & \textbf{Value} \\
\hline
Batch Size & 4 \\
Learning Rate & $1.0 \times 10^{-4}$ \\
Weight Decay & $1.0 \times 10^{-4}$ \\
Dropout Rate & 0.25 \\
Early Stopping Patience & 5 epochs \\
\hline
\end{tabular}
\end{table}

Following the classification benchmark, the validation protocol included a qualitative analysis of motion economy. For this, instrument tip trajectories were generated by plotting the sequence of $(x,y)$ coordinates of the instrument tip keypoint, extracted from the tracking dataset, onto a representative static frame from the corresponding video clip. This method was applied to representative clips from each skill group to enable visual correlation of kinematic data with the skill ratings. 

To conduct the quantitative kinematic analysis, we investigated the relationship between the frame-by-frame tracking data and expert-rated proficiency. We applied the principle of motion economy \cite{40}, which suggests that expert surgeons exhibit greater efficiency than novices. We defined the Cumulative Instrument Path Length ($L$) as the total distance the instrument tip traveled during the recording. Let $P_t = (x_t, y_t)$ represent the two-dimensional coordinates of the instrument tip at frame $t$. The total path length in pixels is calculated as:
\begin{equation}
L = \sum_{t=1}^{T-1} \| P_{t+1} - P_t \|_2 = \sum_{t=1}^{T-1} \sqrt{(x_{t+1} - x_t)^2 + (y_{t+1} - y_t)^2}
\end{equation}
where $T$ is the total number of frames in the sequence where the instrument is visible. This metric serves as a quantitative proxy for surgical efficiency, allowing us to test the hypothesis that the annotated trajectories correlate with the independent, manually adjudicated skill scores.

\section*{Data Records}
All datasets and annotations, including the 3{,}000 raw videos, phase recognition set, instance segmentation set, instrument tracking set, and skill assessment set, are publicly released on Hugging Face under the repository \texttt{mjahmadi/Cataract-LMM}\,\cite{41}.
 
The dataset is structured into five primary directories to facilitate modular access and task-specific downloads. Each primary directory contains a comprehensive \path{README.md} file detailing the specific annotation protocols, data structures, and usage instructions for that subset.

Across all subsets, the filenames for videos and extracted images (e.g., \path{RV_0001_S1.mp4}) include an \path{S1} or \path{S2} identifier to denote the clinical source. This provides crucial metadata for domain adaptation benchmarking: \path{S1} indicates Farabi Eye Hospital, and \path{S2} indicates Noor Eye Hospital.
 
To optimize storage and download efficiency, video files for the four task-specific subsets (1--4) are distributed using a localized archiving strategy. Within the \path{videos/} subdirectory of each respective task, video files are bundled into independent zip archives (e.g., \path{videos_part001.zip}). Each archive is self-contained and filled to an optimal capacity without splitting individual video files across multiple archives. Consequently, users only need to extract the specific archive containing their target video, eliminating the need for multi-part spanned extractions. A \path{ZipContentsReport.csv} file is provided in every task-specific \path{videos/} subdirectory to serve as a directory index, mapping each source video to its corresponding zip archive. In contrast, the 3{,}000 raw recordings in \texttt{5\_Raw\_Videos/videos/} are released as individual \texttt{.mp4} files (e.g., \texttt{RV\_0001\_S1.mp4}) for direct access, accompanied by a \texttt{videos\_metadata.csv} index providing per-video duration, file size, and total frame count. The hierarchical structure of the {Cataract-LMM} repository and the organization of its task-specific subsets are illustrated below:
\begin{center}
\begin{tcolorbox}[
    sharp corners, 
    colback=white, 
    colframe=gray!30, 
    width=0.98\linewidth, 
    boxrule=0.5pt,
    left=15pt
]
\dirtree{%
.1 {\footnotesize \faDatabase} \textbf{Cataract-LMM} \DTcomment{\scriptsize \sffamily Hugging Face Repository}.
.2 {\footnotesize \faFolderOpen} 1\_Phase\_Recognition \DTcomment{\scriptsize \sffamily \color{black!60} Workflow analysis (150 videos + phase segments)}.
.3 {\footnotesize \faFolder} annotations\_full\_video/ \DTcomment{\scriptsize \sffamily \color{black!60} \faFileArchive~Temporal phase labels}.
.3 {\footnotesize \faFolder} annotations\_sub\_clips/ \DTcomment{\scriptsize \sffamily \color{black!60} \faFileArchive~Phase-cut video segments}.
.3 {\footnotesize \faFolder} videos/ \DTcomment{\scriptsize \sffamily \color{black!60} \faFileArchive~Source archives + \faFileCsv~Index report}.
.2 {\footnotesize \faFolderOpen} 2\_Instance\_Segmentation \DTcomment{\scriptsize \sffamily \color{black!60} Scene parsing (6,094 frames)}.
.3 {\footnotesize \faFolder} annotations/ \DTcomment{\scriptsize \sffamily \color{black!60} \faFileCode~COCO JSON \& YOLO TXT formats}.
.3 {\footnotesize \faFolder} videos/ \DTcomment{\scriptsize \sffamily \color{black!60} \faFileArchive~Source archives + \faFileCsv~Index report}.
.2 {\footnotesize \faFolderOpen} 3\_Object\_Tracking \DTcomment{\scriptsize \sffamily \color{black!60} Surgical dynamics (170 clips)}.
.3 {\footnotesize \faFolder} annotations/ \DTcomment{\scriptsize \sffamily \color{black!60} \faFileArchive~Extracted frames \& tracking data}.
.3 {\footnotesize \faFolder} videos/ \DTcomment{\scriptsize \sffamily \color{black!60} \faFileArchive~Source archives + \faFileCsv~Index report}.
.2 {\footnotesize \faFolderOpen} 4\_Skill\_Assessment \DTcomment{\scriptsize \sffamily \color{black!60} Proficiency scoring (170 clips)}.
.3 {\footnotesize \faFolder} annotation/ \DTcomment{\scriptsize \sffamily \color{black!60} \faFileExcel~Objective skill scores}.
.3 {\footnotesize \faFolder} videos/ \DTcomment{\scriptsize \sffamily \color{black!60} \faFileArchive~Source archives + \faFileCsv~Index report}.
.2 {\footnotesize \faFolderOpen} 5\_Raw\_Videos \DTcomment{\scriptsize \sffamily \color{black!60} Complete corpus of surgical recordings}.
.3 {\footnotesize \faFolder} videos/ \DTcomment{\scriptsize \sffamily \color{black!60} \faFileVideo~3,000 raw .mp4 files + \faFileCsv~Metadata index}.
.2 {\footnotesize \faFile*} README.md \DTcomment{\scriptsize \sffamily \color{black!60} Master dataset overview \& instructions}.
}
\end{tcolorbox}
\end{center}
 
The dataset consists of the following five main directories:
 
\begin{enumerate}
    \item \textbf{1\_Phase\_Recognition}: Resources for surgical workflow analysis based on 150 procedures.
    \begin{itemize}
        \item \path{videos/}: Contains the self-contained zip archives of the source videos and the \path{ZipContentsReport.csv} index.
        \item \path{annotations_full_video/}: A zip archive containing individual \path{.csv} files for each source video. Each file provides temporal phase labels structured with the following columns: \path{Video Name}, \path{phase}, \path{start_sec}, \path{end_sec}, \path{start_frame}, and \path{end_frame}.
        \item \path{annotations_sub_clips/}: Contains individual zip archives corresponding to each source video. Each archive is extracted into a directory containing pre-cut video sub-clips, which are automatically segmented at each temporal phase transition from the beginning to the end of the procedure.
    \end{itemize}
 
    \item \textbf{2\_Instance\_Segmentation}: Contains 6{,}094 annotated frames from 150 videos for scene parsing and object segmentation.
    \begin{itemize}
        \item \path{videos/}: Contains the self-contained zip archives of the source videos and the \path{ZipContentsReport.csv} index.
        \item \path{annotations/}: A unified annotations directory that contains both supported label formats as the following two subdirectories:
        \begin{itemize}
            \item \path{annotations/coco_json/}: Provides annotations in the standard COCO format. This includes a single \path{_annotations.coco.json} file and an \path{images/} subdirectory containing an \path{images.zip} file (which should be extracted in place to access the corresponding annotated frames).
            \item \path{annotations/yolo_txt/}: Provides annotations formatted for YOLO architectures. It includes a \path{data.yaml} configuration file, an \path{images/} subdirectory with an \path{images.zip} file, and a \path{labels/} subdirectory with a \path{labels.zip} file. The \path{labels.zip} file contains individual \path{.txt} annotation files corresponding to each image frame. Both zip files are designed to be extracted directly within their respective subdirectories.
        \end{itemize}
    \end{itemize}
 
    \item \textbf{3\_Object\_Tracking}: Contains 170 video clips of the capsulorhexis phase to support spatiotemporal analysis and surgical dynamics.
    \begin{itemize}
        \item \path{videos/}: Contains the self-contained zip archives of the source videos and the \path{ZipContentsReport.csv} index.
        \item \path{annotations/}: Contains individual zip archives for each source video. Each archive includes the extracted visual frames alongside the comprehensive video segmentation tracking annotations and multi-layered data detailed in the tracking methodology.
    \end{itemize}
 
    \item \textbf{4\_Skill\_Assessment}: Provides objective surgical skill ratings for the 170 capsulorhexis video clips.
    \begin{itemize}
        \item \path{videos/}: Contains the self-contained zip archives of the source videos and the \path{ZipContentsReport.csv} index.
        \item \path{annotation/}: Contains a single Excel spreadsheet, 
        \path{skill_scores.xlsx}, that consolidates all multi-indicator skill annotations (e.g., \path{Microscope Use}, \path{Instrument Handling}, \path{Tissue Handling}, \path{Motion}, \path{Commencement of Flap}, \path{Circular Completion}) together with the \path{Adverse Events}, \path{Comment}s, and the final \path{Averaged} proficiency scores per \path{Video_ID}.
        
    \end{itemize}
 
    \item \textbf{5\_Raw\_Videos}: The complete corpus of surgical recordings.
    \begin{itemize}
        \item \path{videos/}: Houses the entirety of the 3{,}000 raw source videos as {individual} \path{.mp4} files, named following the convention \path{RV_<ID>_S<site>.mp4} (e.g., \path{RV_0001_S1.mp4}). A master \path{videos_metadata.csv} index is provided alongside the videos, listing per-file \path{Filename}, \path{Duration (s)}, \path{File Size (MB)}, and \path{Total Frame Count}.
    \end{itemize}
\end{enumerate}

\section*{Technical Validation}
To validate the dataset and demonstrate its utility for multi-task surgical AI, we benchmarked a suite of deep learning models across the four core tasks: phase recognition, instance segmentation, skill assessment, and object tracking. 

\subsection*{Technical Validation on Phase Recognition}
Prior to establishing algorithmic baselines, it is necessary to characterize the intrinsic complexity of the dataset. The phase recognition subset exhibits a pronounced and natural class imbalance, with core phases such as {Phacoemulsification} constituting a substantial portion of the total procedure time, whereas other critical phases, such as {Capsule Polishing}, are significantly shorter, as illustrated in Figure 5. 
\begin{figure}[htbp]
    \centering
    \includegraphics[width=0.7\linewidth]{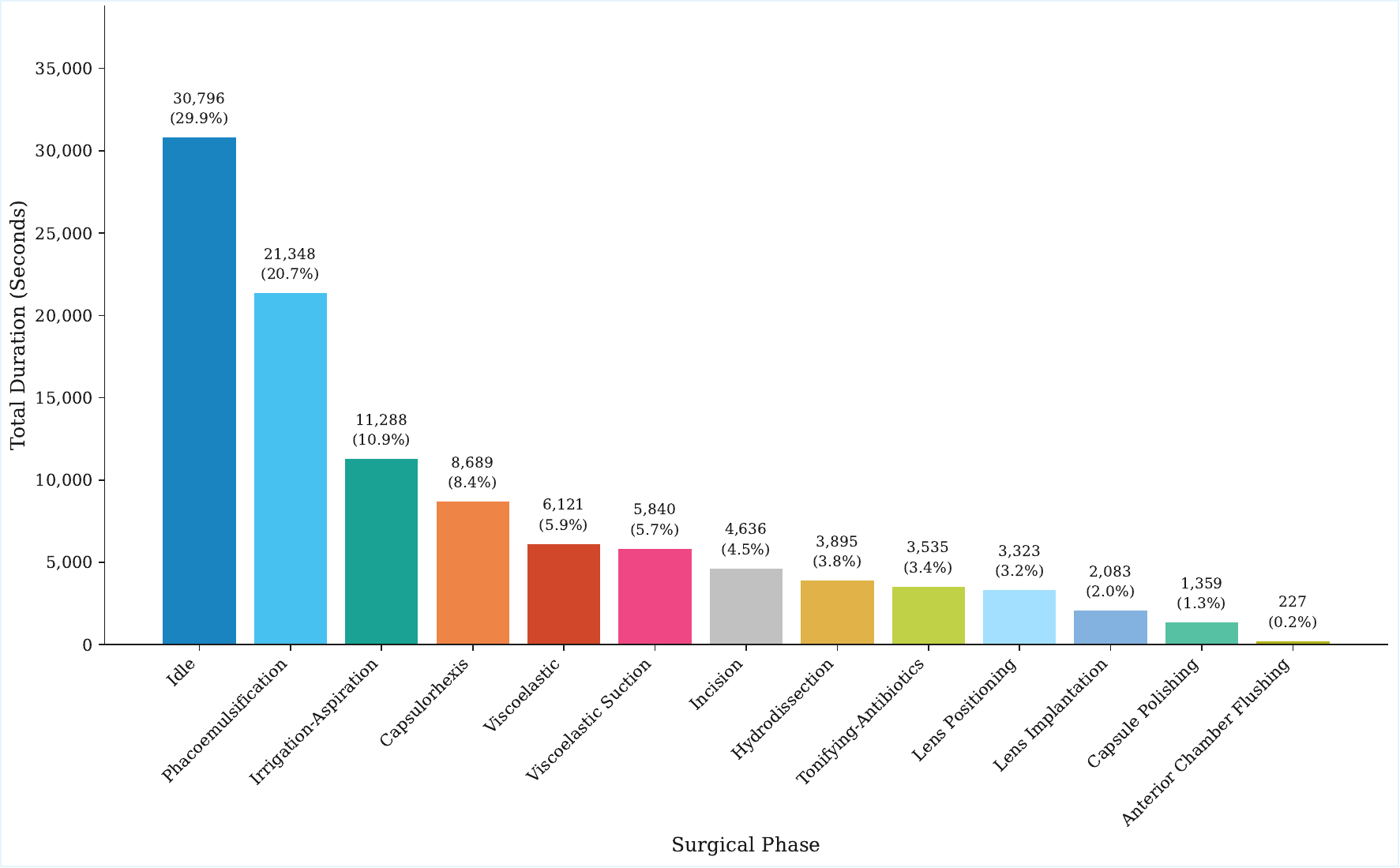}
    \caption{Distribution of total time (in seconds) spent in each surgical phase across the 150 annotated videos.}
    \label{figure5}
\end{figure}

Furthermore, this procedural heterogeneity is further visualized in the normalized timelines of all 150 surgeries (Figure 6). The variations in phase sequence and duration reflect the unscripted nature of the procedures and are attributable to intra-operative events, differing case complexities, and the diverse skill levels of the surgeons. 
\begin{figure}[t]
    \centering
    \includegraphics[width=0.58\linewidth]{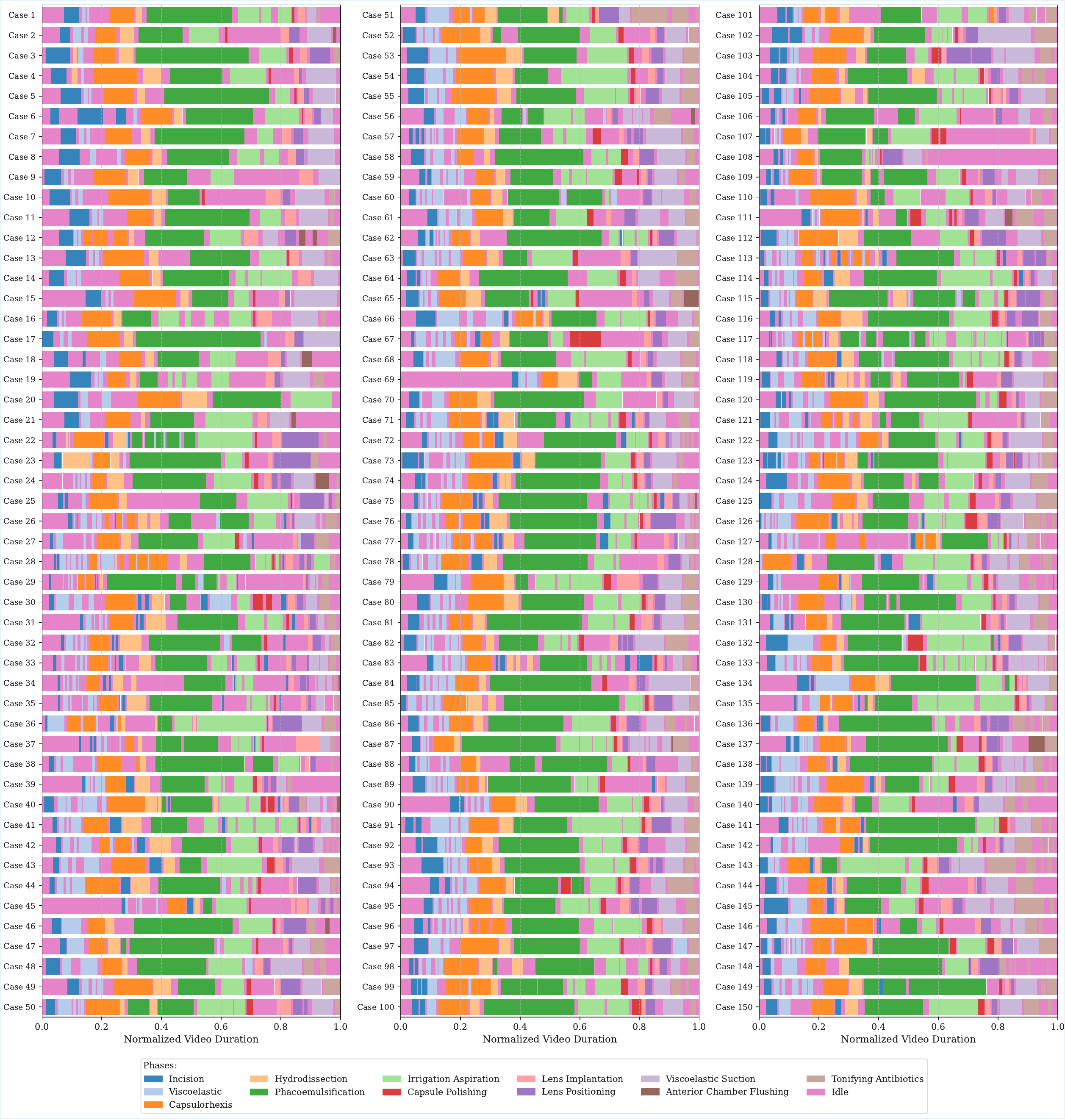}
    \caption{Normalized timelines illustrating procedural heterogeneity across 150 surgeries. Each row represents a single surgery, with phase transitions color-coded, normalized to a standard length from 0 (start) to 1 (end).}
    \label{figure6}
\end{figure}

These inherent variations establish a realistic and challenging baseline for machine learning. To evaluate how well automated systems navigate these complexities and to specifically assess dataset quality under realistic domain shift conditions, the phase recognition annotations were validated through comprehensive benchmarking experiments. Models were trained exclusively on Farabi Hospital videos and evaluated on: (1) 23 unseen Farabi videos (in-domain), and (2) 21 Noor Hospital videos (out-of-domain), using the experimental protocol established in the Methods section.

Table 8 summarizes the performance of clip-level models, which operate on short temporal windows. These models were trained using clips of 10 consecutive frames and are designed for causal (online) prediction, meaning that only past and current frames are available to the model at inference time. On the in-domain (Farabi) test set, Video Transformer architectures showed the highest performance among clip-level approaches, with MViT-B achieving a Macro F1-score of 77.1\%. Hybrid models using an EfficientNet-B5 backbone also achieved strong results (e.g., CNN+GRU, 71.3\% F1-score), while 3D-CNNs such as Slow R50 also performed strongly (69.8\% F1-score). This performance hierarchy validates the dataset’s complexity and its capacity to differentiate architectures based on their spatio-temporal modeling capabilities.
\begin{table}[t]
\centering
\caption{Performance of clip-level models on the in-domain (Farabi) and out-of-domain (Noor) test sets.}
\label{tab8}
\resizebox{\columnwidth}{!}{%
\begin{tabular}{@{}ll *{8}{S[table-format=2.1, table-column-width=1.2cm]}@{}}
\toprule
\multirow{2}{*}{\textbf{Model Architecture}} & \multirow{2}{*}{\textbf{Backbone}} & \multicolumn{4}{c}{\textbf{In-Domain (Farabi Test Set)}} & \multicolumn{4}{c}{\textbf{Out-of-Domain (Noor Test Set)}} \\
\cmidrule(lr){3-6} \cmidrule(lr){7-10}
& & {\shortstack{\textbf{Acc} \\ \textbf{(\%)}}} & {\shortstack{\textbf{F1} \\ \textbf{(\%)}}} & {\shortstack{\textbf{Prec} \\ \textbf{(\%)}}} & {\shortstack{\textbf{Recall} \\ \textbf{(\%)}}} & {\shortstack{\textbf{Acc} \\ \textbf{(\%)}}} & {\shortstack{\textbf{F1} \\ \textbf{(\%)}}} & {\shortstack{\textbf{Prec} \\ \textbf{(\%)}}} & {\shortstack{\textbf{Recall} \\ \textbf{(\%)}}} \\
\midrule
MViT-B & {-} & {\underline{\textbf{85.7}}} & {\underline{\textbf{77.1}}} & 77.1 & {\underline{\textbf{78.5}}} & {\underline{\textbf{71.3}}} & {\underline{\textbf{57.6}}} & 58.5 & {\underline{\textbf{63.1}}} \\
Swin-T & {-} & 85.5 & 76.2 & {\underline{\textbf{77.5}}} & 77.2 & 65.3 & 52.2 & 58.3 & 62.0 \\
\midrule
Slow R50 & {ResNet-50} & 79.6 & 69.8 & 70.7 & 71.3 & 63.4 & 50.5 & {\underline{\textbf{59.3}}} & 59.9 \\
MC3-18 & {ResNet-18} & 78.8 & 67.0 & 71.7 & 69.6 & 51.1 & 43.6 & 55.1 & 50.4 \\
R3D-18 & {ResNet-18} & 74.5 & 64.0 & 67.6 & 66.6 & 47.4 & 41.1 & 56.3 & 51.1 \\
X3D-XS & {-} & 73.3 & 57.1 & 62.3 & 58.7 & 45.9 & 38.3 & 44.6 & 44.1 \\
R(2+1)D-18 & {ResNet-18} & 64.2 & 54.4 & 66.6 & 57.0 & 50.1 & 44.2 & 58.5 & 51.1 \\
\midrule
CNN + GRU & {EfficientNet-B5} & 82.1 & 71.3 & 76.0 & 70.4 & 66.1 & 52.1 & 55.0 & 56.5 \\
% CNN + TeCNO & {EfficientNet-B5} & 81.7 & 71.2 & 75.1 & 71.2 & 64.2 & 49.5 & 55.1 & 53.7 \\
CNN + LSTM & {EfficientNet-B5} & 81.5 & 70.0 & 76.4 & 69.4 & 65.7 & 51.9 & 56.1 & 54.9 \\
CNN + GRU & {ResNet-50} & 79.8 & 69.7 & 70.1 & 70.5 & 43.9 & 42.8 & 54.7 & 48.3 \\
CNN + LSTM & {ResNet-50} & 78.4 & 67.0 & 71.4 & 66.0 & 49.0 & 44.8 & 56.3 & 53.0 \\
% CNN + TeCNO & {ResNet-50} & 77.1 & 66.9 & 68.2 & 69.2 & 46.2 & 41.8 & 49.9 & 53.3 \\
\bottomrule
\end{tabular}%
}
\end{table}

Table 9 presents the results for video-level models, which process entire surgical procedures and therefore leverage non-causal (offline) temporal context, where both past and future frames contribute to predictions. These models incorporate modern temporal architectures and foundation model feature extractors. The ASFormer architecture outperformed earlier temporal variants, highlighting the benefits of long-term context modeling. Notably, the DINO-ASFormer model achieved the highest in-domain F1 score (79.98\%) among evaluated configurations, indicating that DINO, as a self-supervised Vision Transformer, extracts more robust and semantically rich features compared to standard ResNet50 backbones.
\begin{table}[t]
\centering
\caption{Performance of video-level models on the in-domain (Farabi) and out-of-domain (Noor) test sets.}
\label{tab9}
\resizebox{\columnwidth}{!}{%
\begin{tabular}{@{}ll *{16}{c}@{}}
\toprule
\multirow{2}{*}{\textbf{Temporal Model}} & \multirow{2}{*}{\textbf{Backbone}} & \multicolumn{8}{c}{\textbf{In-Domain (Farabi Test Set)}} & \multicolumn{8}{c}{\textbf{Out-of-Domain (Noor Test Set)}} \\
\cmidrule(lr){3-10} \cmidrule(lr){11-18}
& & {\shortstack{\textbf{Acc}}} & {\shortstack{\textbf{F1}}} & {\shortstack{\textbf{Prec}}} & {\shortstack{\textbf{Recall}}} & {\shortstack{\textbf{F1@10}}} & {\shortstack{\textbf{F1@25}}} & {\shortstack{\textbf{F1@50}}} & {\shortstack{\textbf{Edit}}} & {\shortstack{\textbf{Acc}}} & {\shortstack{\textbf{F1}}} & {\shortstack{\textbf{Prec}}} & {\shortstack{\textbf{Recall}}} & {\shortstack{\textbf{F1@10}}} & {\shortstack{\textbf{F1@25}}} & {\shortstack{\textbf{F1@50}}} & {\shortstack{\textbf{Edit}}} \\
\midrule
ASFormer & CLIP & 80.94 & 72.71 & 74.03 & 72.08 & 69.65 & 66.04 & 51.39 & 62.84 & 79.32 & 65.38 & 63.26 & 72.57 & {\underline{\textbf{70.28}}} & {\underline{\textbf{67.32}}} & 56.23 & {\underline{\textbf{60.95}}} \\
ASFormer & DINO & {\underline{\textbf{85.70}}} & {\underline{\textbf{79.98}}} & 80.18 & {\underline{\textbf{80.58}}} & {\underline{\textbf{75.84}}} & {\underline{\textbf{72.99}}} & {\underline{\textbf{62.58}}} & {\underline{\textbf{68.78}}} & {\underline{\textbf{80.64}}} & {\underline{\textbf{67.87}}} & {\underline{\textbf{66.02}}} & {\underline{\textbf{74.50}}} & 68.40 & 64.78 & {\underline{\textbf{57.47}}} & 57.91 \\
ASFormer & ResNet50 & 81.73 & 74.61 & {\underline{\textbf{80.74}}} & 71.46 & 66.64 & 62.50 & 49.93 & 59.56 & 74.26 & 60.93 & 61.33 & 66.38 & 57.08 & 54.16 & 42.69 & 47.32 \\
% TeCNO & CLIP & 72.82 & 59.94 & 63.04 & 58.48 & 55.77 & 49.88 & 36.49 & 50.62 & 75.58 & 59.91 & 59.82 & 63.03 & 61.70 & 57.42 & 48.50 & 52.05
MS-TCN (TeCNO) & CLIP & 76.34 & 67.10 & 69.01 & 67.41 & 58.83 & 53.23 & 41.11 & 53.23 & 76.24 & 62.89 & 61.58 & 68.73 & 62.08 & 57.95 & 48.45 & 51.80
\\
MS-TCN (TeCNO) & DINO & 85.44 & 78.53 & 80.68 & 77.23 & 67.08 & 63.59 & 54.09 & 56.97 & 75.27 & 61.60 & 62.23 & 65.54 & 56.04 & 53.73 & 45.49 & 44.92 \\
MS-TCN (TeCNO) & ResNet50 & 78.50 & 71.75 & 74.60 & 70.19 & 58.77 & 51.75 & 37.62 & 54.60 & 67.32 & 57.52 & 58.87 & 60.69 & 48.78 & 43.03 & 28.91 & 44.86 \\
\bottomrule
\end{tabular}%
}
\end{table}

Evaluation on the out-of-domain (Noor) test set revealed clear differences in model generalization. While traditional fine-tuned models (e.g., MViT-B) experienced an average performance drop of $\sim$22\% due to domain shift (dropping from 77.1\% to 57.6\% F1), models utilizing frozen foundation encoders (DINO and CLIP) showed markedly higher robustness. For instance, the DINO-ASFormer model exhibited a smaller drop of approximately 12\%, achieving 67.9\% F1-score on Noor. Similarly, CLIP-based models maintained stable performance across domains. These results indicate that fine-tuning CNN backbones (such as ResNet50) on data from a single clinic can lead to overfitting to site-specific visual cues (e.g., lighting, microscope texture), whereas employing frozen, pre-trained encoders effectively mitigates this issue, preserving generalization. Collectively, this measurable domain shift highlights a key challenge in surgical AI and reinforces the value of the dataset as a benchmark for developing and evaluating domain adaptation methods.

Furthermore, Figure 7 illustrates the per-phase F1 scores for clip-level and video-level models. Performance patterns are consistent across both modeling levels, revealing a wide performance distribution that validates the dataset’s technical diversity. {Phacoemulsification} is the best-performing phase, which can be attributed to its distinctive instrument and the unique texture of the pupil during this phase. In contrast, Capsule Polishing is the most difficult phase, reflecting its visual similarity to adjacent phases. This marked performance gap between phases demonstrates the dataset's capacity to benchmark a model's sensitivity to fine-grained procedural patterns.
\begin{figure}[htbp]
    \centering
    \includegraphics[width=0.77\columnwidth]{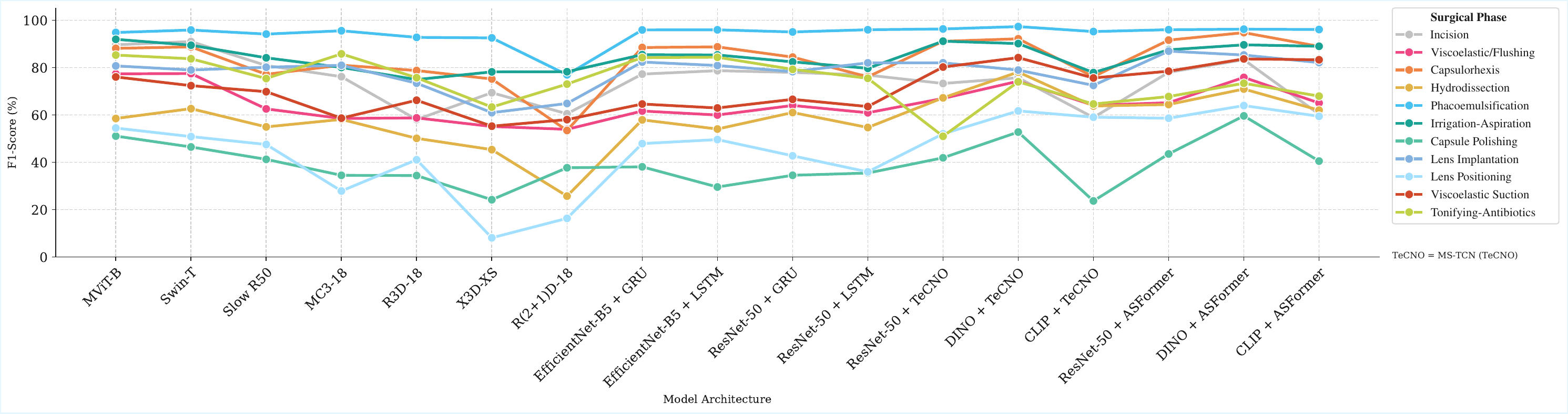}
    \caption{Per-phase F1 scores for all benchmarked models on the in-domain (Farabi) test set.}
    \label{figure7}
\end{figure}

\subsection*{Technical Validation on Instance Segmentation}
Before evaluating model performance, it is critical to characterize the inherent distribution and complexity of the instance segmentation subset. We provide a detailed quantitative analysis of the annotated classes across both clinical centers. Table 10 details the total instance count and sample-level prevalence for each category, quantifying the inherent class imbalance. While anatomical structures appear in nearly all annotated samples, instrument frequency varies according to procedural utility, ranging from common tools like Forceps to rarely used instruments such as the Secondary Knife.
\begin{table}[t]

\centering

\caption{Distribution of annotated instances per class, reported as total count and sample-level prevalence (\%) across the combined dataset and individual clinical centers.}

\label{table10}

\resizebox{0.65 \columnwidth}{!}{%

\begin{tabular}{lllll}

\toprule

\textbf{Category} & \textbf{Class Name} & \textbf{\shortstack{All}} & \textbf{\shortstack{Farabi}} & \textbf{\shortstack{Noor}} \\

\midrule

\multirow{2}{*}{Anatomy} 

 & Pupil & 6,093 (99.98\%) & 3,932 (100\%) & 2,161 (99.95\%) \\

 & Cornea & 6,093 (99.98\%) & 3,932 (100\%) & 2,161 (99.95\%) \\

\midrule

\multirow{10}{*}{Instruments} 

 & Cannula & 1,213 (19.87\%) & 819 (20.78\%) & 394 (18.22\%) \\

 & Capsulorhexis Cystotome & 861 (14.10\%) & 626 (15.90\%) & 235 (10.82\%) \\

 & Capsulorhexis Forceps & 583 (9.57\%) & 451 (11.47\%) & 132 (6.11\%) \\

 & Forceps & 1,923 (31.39\%) & 1,294 (32.68\%) & 629 (29.05\%) \\

 & I/A Handpiece & 920 (14.57\%) & 554 (13.28\%) & 366 (16.93\%) \\

 & Lens Injector & 718 (11.75\%) & 428 (10.86\%) & 290 (13.37\%) \\

 & Phaco Handpiece & 887 (14.54\%) & 512 (13.02\%) & 375 (17.30\%) \\

 & Primary Knife & 756 (12.37\%) & 464 (11.80\%) & 292 (13.41\%) \\

 & Second Instrument & 711 (11.62\%) & 392 (9.92\%) & 319 (14.71\%) \\

 & Secondary Knife & 107 (1.76\%) & 59 (1.50\%) & 48 (2.22\%) \\

\midrule

\textbf{Total Frames} & & \textbf{6,094} & \textbf{3,932} & \textbf{2,162} \\

\bottomrule

\end{tabular}%

}

\end{table}

Furthermore, to quantify the spatial characteristics of the dataset, Table 11 reports the average mask area in pixels for each category. This metric serves as a direct measure of object scale, a critical determinant of segmentation complexity. The wide range of average pixel areas, from large anatomical regions to fine-grained instruments, demonstrates the multi-scale challenges and diversity of this dataset. These metrics also quantify the impact of the heterogeneous acquisition setups, as the higher resolution of the Noor Hospital system ($1920\times1080$) results in significantly larger average mask areas compared to the Farabi Hospital system ($720\times480$).

\begin{table}[t]

\centering

\caption{Average segmentation mask area (in pixels) per instance for each category.}

\label{table11}

\resizebox{0.65 \columnwidth}{!}{%

\begin{tabular}{llSSS}

\toprule

\textbf{Category} & \textbf{Class Name} & {\textbf{All Dataset}} & {\textbf{Farabi (S1)}} & {\textbf{Noor (S2)}} \\

\midrule

\multirow{2}{*}{Anatomy} 

 & Pupil & 50918.30 & 35870.64 & 78285.27 \\

 & Cornea & 112379.11 & 85723.16 & 160857.91 \\

\midrule

\multirow{10}{*}{Instruments} 

 & Cannula & 7029.31 & 4867.33 & 10961.26 \\

 & Capsulorhexis Cystotome & 4318.39 & 4421.08 & 4131.63 \\

 & Capsulorhexis Forceps & 6446.19 & 6330.15 & 6657.22 \\

 & Forceps & 24312.01 & 6903.73 & 55972.17 \\

 & I/A Handpiece & 8336.36 & 5114.39 & 14196.11 \\

 & Lens Injector & 8325.87 & 3722.91 & 16697.23 \\

 & Phaco Handpiece & 15099.51 & 6757.52 & 30270.97 \\

 & Primary Knife & 4004.76 & 3097.11 & 5655.51 \\

 & Second Instrument & 13071.39 & 4285.74 & 29049.73 \\

 & Secondary Knife & 2376.44 & 269.61 & 6208.11 \\

\bottomrule

\end{tabular}%

}

\end{table}

These structural characteristics, specifically the pronounced class imbalance and extreme variations in object scale, create a highly realistic and challenging baseline. To confirm the technical quality of the instance segmentation annotations and assess how models navigate these specific challenges, we performed a series of benchmark experiments on the held-out test set. This validation involved two main analyses: first, a quantitative comparison of supervised models fine-tuned on our dataset against zero-shot architectures to establish baseline performance; and second, an evaluation of the dataset’s utility for tasks requiring different levels of semantic granularity.

Quantitative evaluation of multiple neural network architectures on the 12-class segmentation task is provided in Table 12. The results show that supervised models fine-tuned on our dataset (e.g., YOLOv11-L with 25.3 million parameters, mAP: 73.9) outperform contemporary zero-shot models prompted with ground-truth bounding boxes (e.g., SAM-ViT-H with 632 million parameters, mAP: 56.0). This performance gap validates the quality of the annotations, confirming that the dataset provides the rich, domain-specific signal necessary to train specialized models that exceed the capabilities of general-purpose foundation models on this task. Indeed, prior investigations in this surgical domain have similarly demonstrated that foundation architectures like SAM achieve higher precision when specific components, such as the mask decoder, are explicitly fine-tuned for the visual nuances of the procedure \cite{42}.
\begin{table}[t]
\centering
\caption{Performance benchmark of models on the test set for the 12-class instance segmentation task (Task 3), evaluated using the mAP@0.50:0.95.}
\label{table12}
% S[table-format=2.1] aligns numbers, leaving space for format XX.X
\resizebox{0.65 \columnwidth}{!}{%
\begin{tabular}{@{}l *{5}{S[table-format=2.1]}@{}}
\toprule
\textbf{Class/Model} & {\textbf{Mask R-CNN}} & {\textbf{YOLOv8}} & {\textbf{YOLOv11}} & {\textbf{SAM}} & {\textbf{SAM2}} \\
\midrule
Cornea              & {\underline{\textbf{94.7}}} & 75.9 & 76.3 & 52.7 & 29.7 \\
Pupil               & {\underline{\textbf{91.2}}} & 90.8 & 90.5 & 73.5 & 74.9 \\
Forceps             & 47.0 & 73.8 & {\underline{\textbf{74.5}}} & 48.2 & 58.4 \\
Cannula             & 34.2 & {\underline{\textbf{58.5}}} & 58.4 & 44.5 & 43.2 \\
Phaco Handpiece    & 58.9 & 82.7 & {\underline{\textbf{84.3}}} & 52.4 & 53.8 \\
Second Instrument   & 32.4 & 57.5 & {\underline{\textbf{58.8}}} & 45.7 & 45.2 \\
I/A Handpiece       & 57.9 & 73.9 & {\underline{\textbf{74.8}}} & 50.6 & 54.4 \\
Cap. Cystotome      & 36.8 & {\underline{\textbf{63.1}}} & 62.5 & 44.3 & 42.9 \\
Cap. Forceps        & 15.9 & {\underline{\textbf{66.1}}} & 65.6 & 51.2 & 55.7 \\
Lens Injector       & 36.1 & {\underline{\textbf{84.2}}} & 82.3 & 39.4 & 82.4 \\
Primary Knife       & 79.2 & {\underline{\textbf{89.1}}} & 86.0 & 86.7 & 79.2 \\
Secondary Knife     & 60.2 & 70.9 & {\underline{\textbf{72.0}}} & 39.8 & 62.4 \\
\midrule
All tissue classes     & {\underline{\textbf{92.9}}} & 83.4 & 83.4 & 63.1 & 52.3 \\
All instrument classes  & 45.9 & 71.9 & {\underline{\textbf{72.0}}} & 54.5 & 55.9 \\
Overall (All Classes)             & 53.7 & 73.8 & {\underline{\textbf{73.9}}} & 56.0 & 55.2 \\
\bottomrule
\end{tabular}
}
\end{table}

A per-class analysis reveals that segmenting anatomical structures (e.g., Pupil, mAP: 90.5) is a less difficult task than segmenting instruments, which are subject to visual challenges such as motion blur, specular reflections, and fine structural details. The lower performance on thin instruments (e.g., Cannula, mAP: 58.4) underscores the challenging and realistic nature of the dataset. The qualitative comparison in Figure 8 visually confirms the higher precision of the fine-tuned supervised model.
\begin{figure}[htbp]
    \centering
    \includegraphics[width=0.65\linewidth]{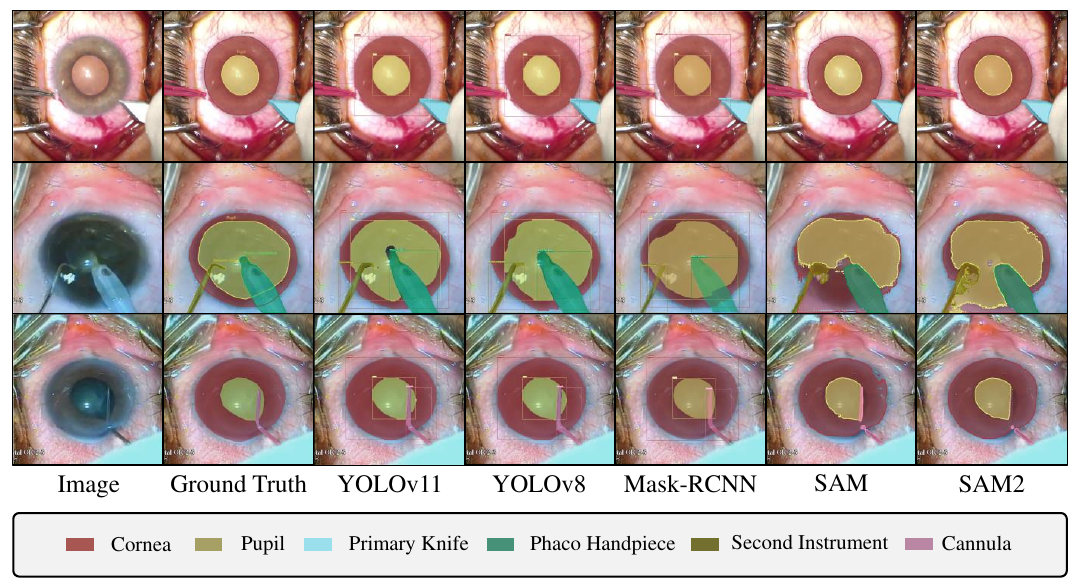}
    \caption{Qualitative comparison of segmentation outputs on task 2.}
    \label{fig10}
\end{figure}

To assess the dataset's flexibility for different applications, we evaluated the performance of the top-performing model, YOLOv11-L, across three tasks with varying class granularity. The results, detailed in Table 13, demonstrate a clear trade-off between semantic detail and segmentation accuracy:
\begin{enumerate}
    \item Task 1 (3 Classes): By consolidating all instruments into a single ‘Instrument’ class, the model effectively mitigates class confusion, achieving a high mask mAP of 74.0 for this unified class. This demonstrates the dataset's utility for high-level tasks where only instrument presence detection is required.

    \item Task 2 (9 Classes): This intermediate task, which merges only the most visually similar instruments, yielded the highest overall mask mAP of 75.17. This balanced approach reduces ambiguity while retaining significant detail, validating the dataset for robust, multi-class instrument recognition.

    \item Task 3 (12 Classes): While the most challenging due to high inter-class similarity, this task still yielded a strong overall mAP of 73.19. This result confirms that the dataset contains sufficient distinguishing visual features to train models for fine-grained analysis, where distinguishing specific instruments is critical.
    
\end{enumerate}

\begin{table}[t]
\centering
\caption{YOLOv11-L model performance (mAP@0.50:0.95) on the test set across three segmentation tasks with varying semantic granularity.}
\label{tab10}
% S[table-format=2.2] aligns numbers and sets space for format XX.XX
\resizebox{0.55\textwidth}{!}{%
\begin{tabular}{@{}l S[table-format=2.2] S[table-format=2.2] S[table-format=2.2]@{}}
\toprule
\textbf{Class} & {\textbf{\shortstack{Task 1 \\ (3 classes)}}} & {\textbf{\shortstack{Task 2 \\ (9 classes)}}} & {\textbf{\shortstack{Task 3 \\ (12 classes)}}} \\
\midrule
Cornea                     & 75.4  & 75.5  & 75.7  \\
Pupil                      & 91    & 90.1  & 90.4  \\
Instrument (All)           & 74  & {---} & {---} \\
Instrument (Grouped)       & {---} & 60.7  & {---} \\
Knife (Grouped)            & {---} & 81.1  & {---} \\
Capsulorhexis Forceps      & {---} & 60.1  & 63.4  \\
Forceps                    & {---} & 70.9  & 73.0  \\
Lens Injector              & {---} & 80.6  & 81    \\
Phaco Handpiece           & {---} & 83.6  & 82.8  \\
I/A Handpiece              & {---} & 74.0  & 73.5  \\
Primary Knife              & {---} & {---} & 85.2  \\
Secondary Knife            & {---} & {---} & 77.9  \\
Capsulorhexis Cystotome    & {---} & {---} & 61.9  \\
Second Instrument          & {---} & {---} & 57.4  \\
Cannula                    & {---} & {---} & 56.1  \\
\midrule
All tissue classes         & 83.2  & 82.8  & 83.05 \\
All Instrument classes     & 74    & 73    & 71.22 \\
Overall (All Classes)      & 80.13  & 75.17  & 73.19  \\
\bottomrule
\end{tabular}
}
\end{table}

To quantify domain shift, we conducted a cross-center validation study for instance segmentation. Retaining the original hyperparameters and modifying only the source-based split, we trained the top-performing YOLOv11-L model independently on subsets from each surgical center (Farabi and Noor) and evaluated performance on in-distribution and out-of-distribution test sets. This analysis, conducted on the 12-class instance segmentation task (Task 3), captures the impact of varying hardware specifications, distinct surgical instrument sets, and diverse visual appearances.

As detailed in Table 14, experimental results demonstrate a performance degradation in cross-center scenarios. Specifically, the model trained on Farabi Hospital (S1) data achieved an overall mask mAP (0.50:0.95) of 70.45\%, which declined to 61.19\% when evaluated on Noor Hospital (S2). A larger shift occurred in the inverse scenario, where performance dropped from 61.99\% to 47.12\%. Comparative analysis indicates that anatomical structures (Cornea, Pupil) are more robust to domain shift than specialized surgical instruments. For instance, the {Phaco Handpiece} exhibited high sensitivity to domain shift, with a performance drop of 21.3\% in the inverse transfer setting. This degradation is likely attributable to distinct cross-center variations in the instrument's visual appearance. This gap underscores the challenge of heterogeneous acquisition environments, which introduce variations in resolution and visual appearance, highlighting the dataset's potential for benchmarking domain-adaptive models.
\begin{table}[t]
\centering
\caption{Quantitative evaluation of cross-center domain shift for the 12-class instance segmentation task (Task 3). Performance is reported as mAP@50:95 for the YOLOv11-L model trained and tested on distinct clinical centers (Farabi vs. Noor).}
\label{tab132}
\resizebox{0.77 \columnwidth}{!}{%
\begin{tabular}{llcccc}
\toprule
\multirow{2}{*}{\textbf{Category}} & \multirow{2}{*}{\textbf{Class Name}} & \multicolumn{2}{c}{\textbf{Train: Farabi}} & \multicolumn{2}{c}{\textbf{Train: Noor}} \\
\cmidrule(lr){3-4} \cmidrule(lr){5-6}
 & & \textbf{Test: Farabi} & \textbf{Test: Noor} & \textbf{Test: Noor} & \textbf{Test: Farabi} \\
\midrule
\multirow{2}{*}{Anatomy} 
 & Cornea & 70.4 & 66.9 & 67.5 & 59.5 \\
 & Pupil & 86.7 & 83.1 & 72.1 & 65.4 \\
\midrule
\multirow{10}{*}{Instruments} 
 & Forceps & 69.7 & 60.2 & 61.9 & 46.6 \\
 & Cannula & 61.5 & 50.9 & 50.4 & 37.22 \\
 & Phaco Handpiece & 81.1 & 70.8 & 72.8 & 51.5 \\
 & Second Instrument & 55.3 & 44.0 & 48.5 & 30.02 \\
 & I/A Handpiece & 71.5 & 60.1 & 60.0 & 39.17 \\
 & Capsulorhexis Cystotome & 60.4 & 49.4 & 56.0 & 41.89 \\
 & Capsulorhexis Forceps & 62.4 & 53.5 & 54.65 & 41.69 \\
 & Lens Injector & 75.1 & 68.3 & 64.2 & 46.2 \\
 & Primary Knife & 84.9 & 75.0 & 77.0 & 59.0 \\
 & Secondary Knife & 66.4 & 52.0 & 58.9 & 47.3 \\
\midrule
\multicolumn{2}{l}{\textbf{Overall (All Classes)}} & \textbf{70.45} & \textbf{61.19} & \textbf{61.99} & \textbf{47.12} \\
\bottomrule
\end{tabular}
}
\end{table}

\subsection*{Technical Validation on Object Tracking}
To evaluate the technical quality of the object tracking annotations, we performed a quantitative analysis focusing on two dimensions: spatial distribution and interaction proximity.

First, to evaluate spatial consistency, we analyzed the surgical workspace dynamics by comparing the spatial envelopes of the Lower-Skilled and Higher-Skilled cohorts. The resulting probability maps (Figure 9) delineate a concentrated Region of Interest for the capsulorhexis phase, while revealing distinct kinematic patterns associated with surgical proficiency.
\begin{figure}[htbp]
    \centering
    \includegraphics[width=0.8\linewidth]{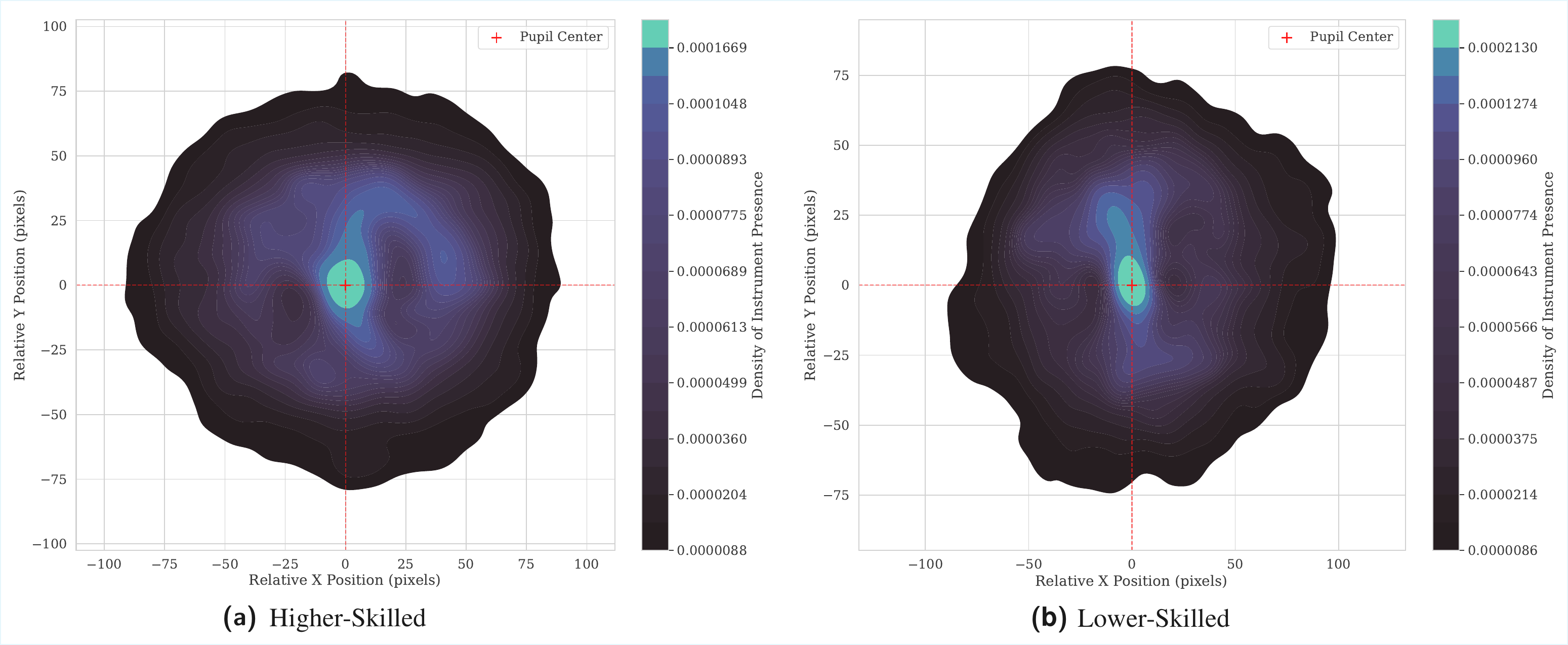}
    \caption{Spatial consistency analysis of instrument tracking annotations using 2D Kernel Density Estimation (KDE).}
    \label{figure9}
\end{figure}

The distribution is densely concentrated within a $\pm$100-pixel radius of the pupil centroid $(0,0)$ for both groups, indicating that the annotated trajectories consistently adhere to the expected anatomical constraints of the procedure. However, the Lower-Skilled cohort exhibits a more tightly localized, vertically elongated concentration with a higher peak probability density of approximately $2.13 \times 10^{-4}$. 

In contrast, the Higher-Skilled cohort demonstrates a broader and more uniformly distributed spatial envelope around the center, with a lower peak density of approximately $1.67 \times 10^{-4}$. These peak values represent the probability density per unit area, indicating an approximate likelihood of $\sim$0.021\% and $\sim$0.017\%, respectively, of instrument presence at the exact central pixel in any given frame.

This morphological difference quantitatively captures the smooth, continuous circular sweeping motion characteristic of expert capsulorhexis, distinguishing it from the more repetitive, constrained, or hesitant central interactions typical of novices. Furthermore, the absence of significant density artifacts in the periphery ($>150$ pixels) across both maps indicates the reliability of the tracking annotations against false positives in the background.

Following the spatial envelope analysis, the overall distribution of the Euclidean distance between the instrument tip and the pupil centroid (Figure 10) exhibits a unimodal form with a pronounced positive skew. This characterizes the capsulorhexis as a highly centralized procedure, with a median interaction distance of 43.0 pixels and a mean of 45.6 pixels. The tight clustering of the KDE curve around this central tendency further supports the consistency of the surgical ROI captured by the annotations. The extended right-sided tail corresponds to distinct, transient procedural events such as instrument insertion and withdrawal at the peripheral corneal incision, indicating that the dataset captures the full range of surgical kinematics.
\begin{figure}[htbp]
    \centering
    \includegraphics[width=0.7\linewidth]{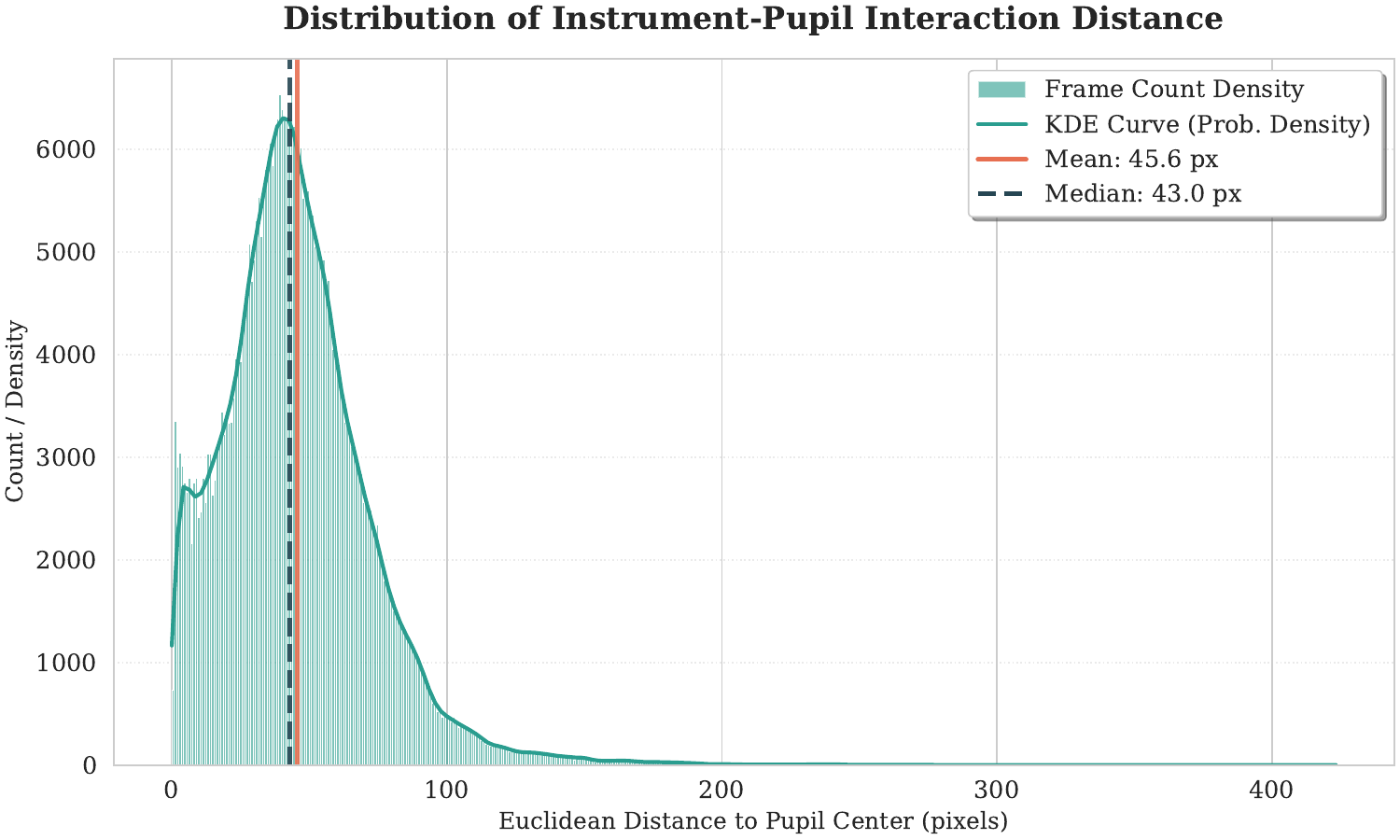}
    \caption{Distribution of instrument-to-pupil Euclidean distances.}
    \label{fig11_2}
\end{figure}

Beyond the geometric and kinematic characterization of the labels, the tracking annotations provided in this subset can be utilized to evaluate and train various computer vision models. For methodologies such as Multi-Object Tracking (MOT), the paired bounding boxes and persistent instance identifiers provide data required to optimize modular Tracking-by-Detection frameworks. This includes calibrating association algorithms like ByteTrack \cite{43} and training robust detectors like the YOLO series \cite{44}. Furthermore, this annotated data supports end-to-end approaches, including query-based transformers like MOTR \cite{45}, alongside the Siamese-based single-object trackers that are effective for capsulorhexis instrument tracking \cite{46}.

Finally, expanding beyond bounding box tracking into the domains of Video Instance Segmentation (VIS) and Multi-Object Tracking and Segmentation (MOTS), the detailed pixel-level instance masks included in the dataset can be applied to the training of Masked-Attention Transformers such as Mask2Former \cite{47} and surgical variants like MATIS \cite{48}. They also serve as highly accurate prompts for the few-shot adaptation of foundation models, specifically TAM \cite{49}. Additionally, point-tracking transformers such as CoTracker \cite{50} can effectively leverage the dense keypoint labels to model robust motion dynamics entirely independent of rigid body assumptions, ensuring comprehensive surgical scene understanding.

\subsection*{Technical Validation on Skill Assessment}
To evaluate the skill assessment annotations, we first analyzed the distribution and construct validity of the ground-truth data. Analysis of the aggregated overall scores shows a continuous distribution of surgical skill. The composite visualization in Figure 11 details this distribution for all 170 rated clips. The histogram illustrates the frequency of scores, which approximates a normal distribution with a slight negative skew (skewness = -0.31).

The accompanying box plot provides summary statistics, showing a median score of 3.85, an interquartile range (IQR) from 3.39 to 4.36, and a total range from 2.29 to 5.00. This well-characterized distribution provides the data necessary to benchmark skill assessment models. To define the classes used in the benchmarking experiments, the plot background is shaded to distinguish between the {Lower-skilled} and {Higher-skilled} groups identified via K-Means clustering (refer to the Experimental Methodology for Skill Assessment section).

Furthermore, the overlay of average procedural duration (orange line) highlights an inverse relationship between skill scores and surgical time. As skill scores increase, procedural duration generally decreases; for instance, clips in the lowest score range ([2.29 - 2.50]) average 137.77s ($\pm$48.82s), whereas those in the highest range ([4.79 - 5.00]) average just 26.30s ($\pm$9.68s). 
\begin{figure}[htbp]
    \centering
    \includegraphics[width=1\linewidth]{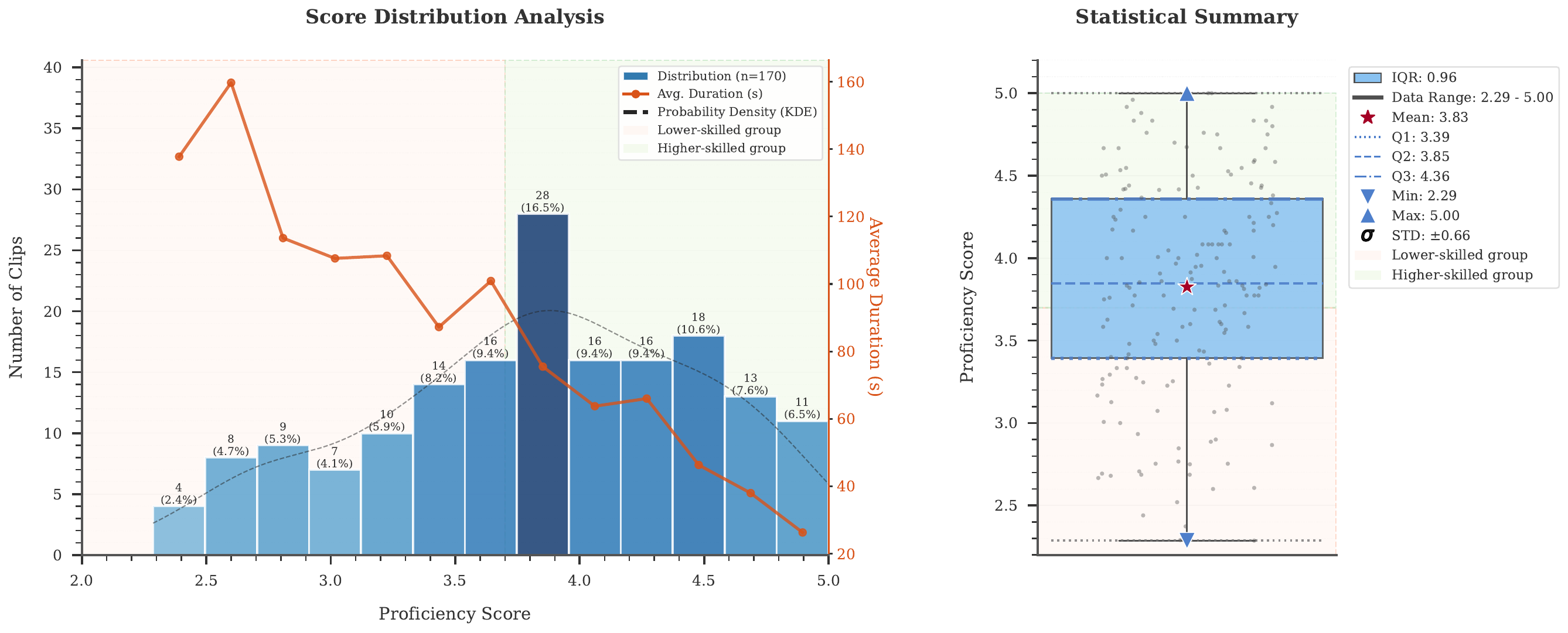}
    \caption{Skill score distribution overlaid with average procedural duration and a statistical box plot. Shaded regions denote the skill groups used for benchmarking.}
    \label{figure11}
\end{figure}

Following the distribution analysis, a Pearson correlation analysis was performed between the six performance indicators and the procedural duration to assess the construct validity of the rubric. The heatmap in Figure 12 details this analysis, revealing positive correlations between core psychomotor domains, such as {Instrument Handling} and {Motion} (r=0.74), and between {Motion} and {Circular Completion} (r=0.78). This indicates that the rubric captures distinct but related facets of surgical technique. Furthermore, all six performance indicators were negatively correlated with procedural duration.
\begin{figure}[t]
    \centering
    \includegraphics[width=0.65\linewidth]{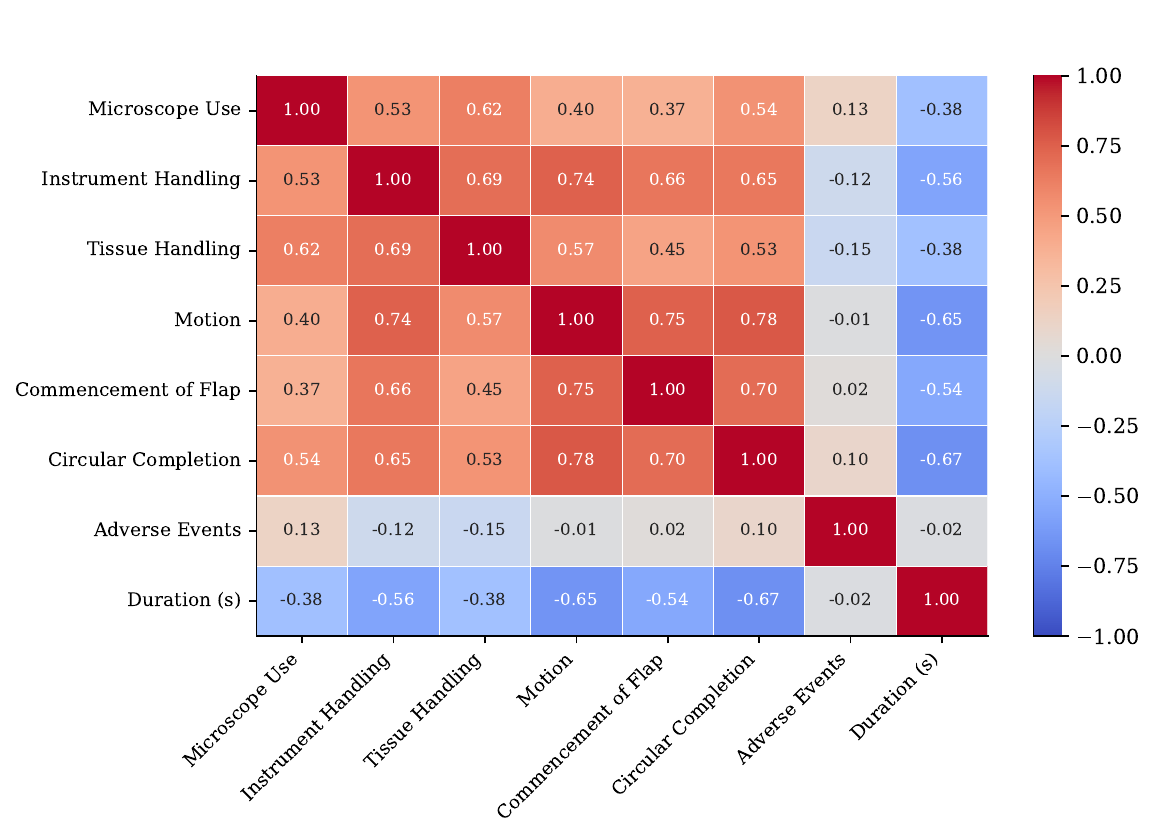}
    \caption{Pearson correlation matrix for the six skill assessment indicators and procedural duration.}
    \label{figure12}
\end{figure}

Using these defined skill groups, a video-based classification benchmark was performed on a held-out test set using the binary skill groups ({Lower-Skilled} and {Higher-Skilled}) defined previously. The resulting performance metrics are detailed in Table 15.

Among the benchmarked models, TimeSformer achieved the highest F1-score (83.90\%). This result indicates that the dataset's skill labels correlate with visual features, providing a baseline for automated assessment classification. While 3D-CNNs also performed well (e.g., R3D-18 F1-score: 83.58\%), the lower performance of hybrid CNN-RNN models (e.g., CNN-GRU F1-score: 68.57\%) indicates that long-range spatiotemporal feature extraction is more effective for modeling these skill classifications.

\begin{table}[b]
\centering
\caption{Performance comparison of various video classification models on the test set.}
\label{table15}
% The 'S' column type from the siunitx package is used for alignment.
\begin{tabular}{@{}l *{4}{S[table-format=2.2]}@{}}
\toprule
\textbf{Model} & {\textbf{Accuracy (\%)}} & {\textbf{Precision (\%)}} & {\textbf{Recall (\%)}} & {\textbf{F1-Score (\%)}} \\
\midrule
TimeSformer & {\underline{\textbf{82.50}}} & {\underline{\textbf{86.00}}} & 82.00 & {\underline{\textbf{83.90}}} \\
R3D-18      & 81.67 & 82.35 & {\underline{\textbf{84.85}}} & 83.58 \\
SlowFast R50    & 80.00 & 81.82 & 81.82 & 81.82 \\
X3D-M       & 80.00 & 83.87 & 78.79 & 81.25 \\
R(2+1)D-18  & 72.92 & 79.31 & 76.67 & 77.97 \\
CNN-LSTM    & 61.67 & 70.97 & 66.67 & 68.75 \\
CNN-GRU     & 54.17 & 60.00 & 80.00 & 68.57 \\
\bottomrule
\end{tabular}
\end{table}

In addition to video-level classification, the tracking annotations permit a detailed geometric analysis of surgical motion. Figure 13 presents a qualitative comparison of instrument tip trajectories for two surgeons with contrasting proficiency scores. The trajectory of the highly-rated expert surgeon (Figure 13a) is characterized by a smooth, continuous, and circular path that minimizes deviations from the intended capsulorhexis boundary. In distinct contrast, the lower-rated surgeon's trajectory (Figure 13b) is marked by high-frequency jitter, frequent direction changes, and erratic backtracking. These differences indicate that the tracking annotations capture kinematic patterns associated with surgical dexterity.
\begin{figure}[b]
    \centering
    \includegraphics[width=0.85\linewidth]{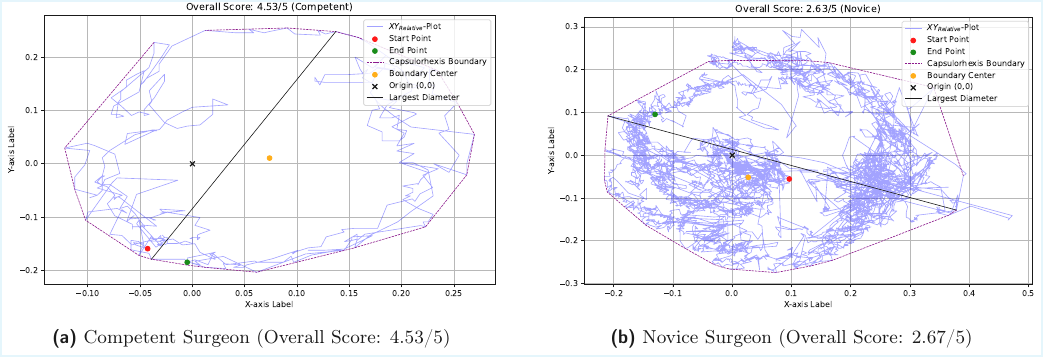}
    \caption{Instrument tip trajectories during the capsulorhexis phase, visualizing the difference in motion economy between an expert and a novice surgeon.}
    \label{figure13}
\end{figure}

To translate these qualitative observations into objective metrics, we analyzed the correlation between the derived {Cumulative Instrument Path Length} and the expert skill scores. As illustrated in Figure 14, linear regression analysis shows an inverse relationship between kinematic path length and surgical proficiency. To facilitate granular retrospective analysis, each data point in the plot is embedded with the unique video identifier corresponding to the specific clip in the tracking/skill subset.

The regression line exhibits a negative slope of $-5,803.8$, indicating that for every single-point increase in the expert skill score, the total instrument travel distance decreases by approximately 5,800 pixels. Stratification by skill level reveals distinct kinematic signatures: the Lower-Skilled cohort (Score $< 3.7$) exhibits a significantly higher mean path length of $13,724.6$ pixels, reflecting tentative and inefficient movements, whereas the Higher-Skilled cohort (Score $\ge 3.7$) demonstrates a markedly reduced mean path length of $7,124.7$ pixels. This nearly twofold reduction indicates that path length derived from the tracking annotations correlates with the expert skill scores.
\begin{figure}[htbp]
    \centering
    \includegraphics[width=0.75\linewidth]{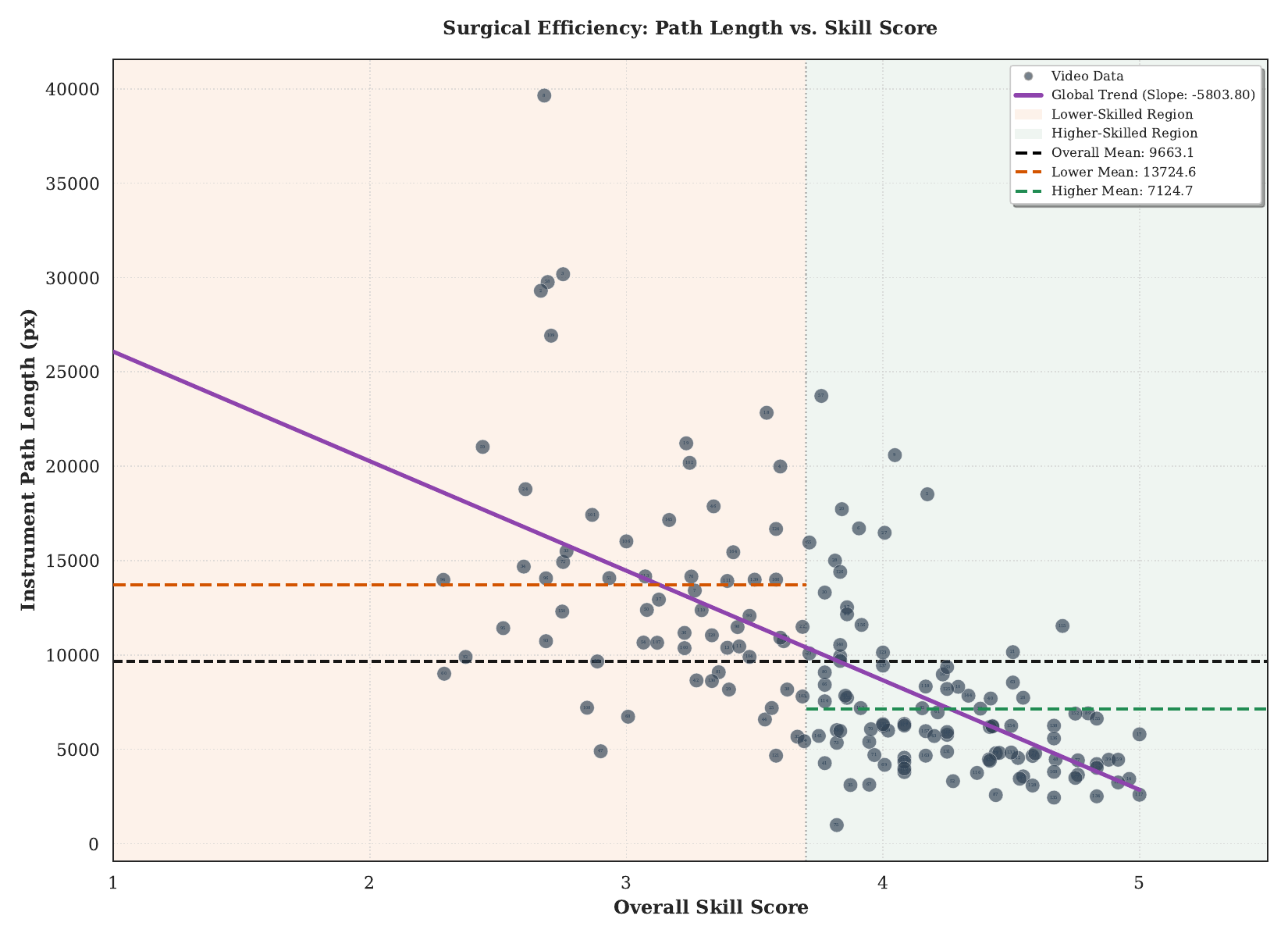}
    \caption{Correlation between cumulative instrument path length and surgical proficiency.}
    \label{figure14}
\end{figure}

Beyond classification and initial kinematic analysis, the labels in the Cataract-LMM skill assessment subset facilitate the development of diverse action quality assessment pipelines. The stratified skill groupings provide the necessary supervision for {Surgical Skill Classification}, supporting the optimization of spatiotemporal networks such as the CNN-RNN architectures \cite{51}, and 3D CNNs \cite{52}. Furthermore, the granular rubric scores enable {Continuous Skill Regression}, serving as ground truth for hybrid CNN-BiLSTM frameworks like ViSA \cite{53}. 

Finally, combining these skill scores with the tracking data validates metric-based assessment pipelines, allowing geometric and kinematic descriptors to be correlated with standardized competency ratings, comparable to PhacoTrainer \cite{54}. While our path length analysis demonstrates a baseline for this approach, the dense temporal resolution of the Cataract-LMM tracking labels supports the development of more advanced kinematic indicators, such as average velocity or jerk-based metrics \cite{41}.

Beyond the detailed validation of the annotated subsets, the complete corpus of 3,000 videos can be utilized for large-scale representation learning. In the context of foundation models, this volume of unannotated data supports self-supervised learning (SSL) paradigms. Specifically, the dataset can be applied to image-level contrastive learning frameworks, such as MoCo v2 \cite{541}, to extract domain-invariant features. Research by Ramesh et al. \cite{55} demonstrates that pretraining deep learning backbones on domain-specific ophthalmic data yields measurable improvements over standard ImageNet initialization. Their findings indicate that initializing models with SSL weights learned on cataract surgery data (CATARACTS dataset) yields performance boosts of $\sim$1–4\% over ImageNet initialization.

Extending these representation paradigms to the temporal domain, frameworks such as SurgVISTA \cite{56} utilize unannotated video sequences to model long-range dependencies without dense supervision. Empirical validation in the ophthalmic domain indicates that pretraining on large-scale surgical video enables SurgVISTA to outperform image-based baselines, with reported improvements of 14.3\% in phase-level Jaccard indices on the CATARACTS dataset. The unannotated portion of Cataract-LMM can therefore be utilized to investigate similar data-scaling approaches, providing the video sequences required to extract the generalizable spatiotemporal features necessary for analyzing complex surgical workflows.

Alongside these purely visual approaches, the dataset can also be incorporated into multimodal research. While it provides a large volume of visual data, a notable limitation is the absence of intrinsic textual reports or metadata for the raw videos. However, this constraint does not preclude the development of Vision-Language Pre-training (VLP) models. Recent hierarchical retrieval-augmented frameworks, such as OphCLIP \cite{57}, address this by leveraging silent surgical videos as a knowledge base to enhance representation learning. By aligning visual representations with structured text prompts, a methodology validated on the Cataract-1K and Cataract-101 datasets, these methods enable zero-shot phase recognition and the generation of fine-grained attention maps for interpretability, effectively localizing instruments and anatomical structures without pixel-level supervision. Consequently, the dataset can be used as a resource for evaluating label-efficient understanding even in the absence of paired text.

Furthermore, beyond representation and multimodal learning, the technical heterogeneity of the dataset, which encompasses multi-center data with varying acquisition protocols, provides data for Generative AI applications. This diversity can be applied to style transfer frameworks like SurReal \cite{58}, which can leverage the raw footage to bridge the simulation-to-clinical gap, facilitating the adaptation of VR-trained models to real-world operative environments. Additionally, the number of procedures provides material for the training of controllable generative video models such as SurgSora \cite{59}. These models can synthesize privacy-preserving datasets and generate realistic instances of rare, safety-critical adverse events, which are methods used to mitigate class imbalance and evaluate the robustness of surgical AI systems.

\section*{Data Availability}
The Cataract-LMM dataset supporting this Data Descriptor is publicly available on Hugging Face under the repository \texttt{mjahmadi/Cataract-LMM}\,\cite{41}.

To optimize storage and modular access, the deposit is organized into five primary directories, each containing a comprehensive \path{ReadMe.md} file and localized video zip archives indexed by a \path{ZipContentsReport.csv} file. Specifically, the repository contains: (i) \path{1_Phase_Recognition} resources, including full-video temporal annotations (.csv) and pre-cut phase sub-clips; (ii) \path{2_Instance_Segmentation} annotations provided in both COCO (.json) and YOLO (.txt) formats with their corresponding extracted image frames; (iii) \path{3_Object_Tracking} multi-layered tracking data with per-frame annotations (.json) and extracted frames; (iv) \path{4_Skill_Assessment} containing the multi-indicator objective proficiency scores (.xlsx); and (v) \path{5_Raw_Videos} housing the complete corpus of 3,000 unannotated surgical recordings (.mp4) accompanied by a \path{videos_metadata.csv} index.

\section*{Usage Notes}
The Cataract-LMM dataset is licensed under the Creative Commons Attribution 4.0 International (CC-BY 4.0) license. This permits unrestricted access, use, and redistribution of the data, provided that appropriate credit is given. We respectfully request that users cite this publication in any research or derivative works utilizing the dataset. For community support, detailed tutorials, and future updates, please refer to the project's \href{https://mjahmadee.github.io/Cataract-LMM/}{website} (\url{https://mjahmadee.github.io/Cataract-LMM/}) and \href{https://github.com/MJAHMADEE/Cataract-LMM}{GitHub repository} (\url{https://github.com/MJAHMADEE/Cataract-LMM}).

\section*{Code Availability}
The source code utilized for data preprocessing, along with the complete implementation used to generate all baseline results reported in the Technical Validation section, is publicly accessible in the project's \href{https://mjahmadee.github.io/Cataract-LMM/}{website} (\url{https://mjahmadee.github.io/Cataract-LMM/}) and \href{https://github.com/MJAHMADEE/Cataract-LMM}{GitHub repository} (\url{https://github.com/MJAHMADEE/Cataract-LMM}).

\section*{Author contributions}
M.J.A. and H.D.T. jointly conceptualized the study. M.J.A. also designed the methodology, performed the primary data analysis and technical validations, managed the project, and wrote the original manuscript.
P.A., S.F.M., and M.K. assisted with clinical data acquisition, curation, and validation at Farabi Hospital, while S.F.M. and H.H. performed similar roles at Noor Hospital. 
Additionally, M.J.A., I.G., and P.A. collaboratively managed the data collection protocols and engineered the associated data processing pipelines. 
I.G. and A.T. contributed to the technical validation of the instance segmentation and phase recognition tasks, respectively. 
Final supervision and critical review of the project were conducted from two complementary perspectives. P.A. and S.F.M. provided the medical validation, ensuring the clinical accuracy and relevance of the datasets, results, and the final manuscript. Concurrently, H.D.T. and M.T. provided the technical and engineering (AI-related) validation and supervision, critically reviewing the methodologies, computational results, and the manuscript from an engineering standpoint. Finally, all authors reviewed and approved the final manuscript.

\section*{Competing interests}
The author(s) declare no competing interests.

\end{document}